\documentstyle[twoside,palatino,cl,epsfig]{article}

\let\cbstart\relax
\let\cbend\relax

\issue{26}{3}{2000}

\begin{document}

\bibliographystyle{fullname}

\title{Dialogue Act Modeling for
\mbox{Automatic Tagging and Recognition}
\mbox{of Conversational Speech}}

\runningauthor{Stolcke et al.}
\runningtitle{Dialogue Act Modeling}

\author{Andreas Stolcke%
         \thanks{Speech Technology and Research Laboratory, SRI International,
		333 Ravenswood Ave., Menlo Park, CA 94025,
		1-650-859-2544. E-mail: stolcke@speech.sri.com.}
       					& Klaus Ries \\
	\affil{SRI International}       & \affil{Carnegie Mellon University and} \\
					& \affil{University of Karlsruhe} \\
		\\
	Noah Coccaro                    & Elizabeth Shriberg    \\
	\affil{University of Colorado at Boulder} & \affil{SRI International} \\
		\\
	Rebecca Bates			& Daniel Jurafsky	\\
	\affil{University of Washington}& \affil{University of Colorado at Boulder} \\
		\\
	Paul Taylor			& Rachel Martin \\
	\affil{University of Edinburgh}	& \affil{Johns Hopkins University} \\
		\\
	Carol Van Ess-Dykema 		& Marie Meteer \\
	\affil{U.S. Department of Defense} & \affil{BBN Technologies} \\
}

\maketitle 



\newcommand{\argmax}{\mathop{\rm argmax}}
\newcommand{\cstart}{\mbox {\tt <c>}}
\newcommand{\B}[0]{$\bullet$}

\newcommand{\DA}[1]{{\sc #1}}

\renewcommand{\textfraction}{0.1}
\renewcommand{\floatpagefraction}{0.9}

\begin{abstract}
We describe a statistical approach for modeling dialogue acts in
conversational speech, i.e.,  speech-act-like
units such as \DA{Statement}, \DA{Question}, \DA{Backchannel}, \DA{Agreement},
\DA{Disagreement}, and \DA{Apology}. Our model detects and predicts dialogue acts
based on lexical, collocational, and prosodic cues, as well as on the
discourse coherence of the dialogue act sequence.  
The dialogue model is based on treating the discourse structure of
a conversation as a hidden Markov model and the individual dialogue acts
as observations emanating from the model states.  Constraints on the likely
sequence of dialogue acts are modeled via a dialogue act $n$-gram.  
\cbstart
The statistical dialogue grammar is combined with word $n$-grams,
decision trees, and neural networks modeling the idiosyncratic
lexical and prosodic manifestations of each dialogue act. 
We develop a probabilistic integration of speech recognition with
dialogue modeling, to improve both speech recognition and dialogue act
classification accuracy.
Models are trained and evaluated using a large hand-labeled database of
1,155~conversations from the Switchboard corpus of spontaneous
\cbend
human-to-human telephone speech.  We achieved good dialogue act labeling
accuracy (65\% based on errorful, automatically recognized words and
prosody, and 71\% based on word transcripts, compared to a chance baseline
accuracy of 35\% and human accuracy of 84\%) and a small reduction in
word recognition error.
\end{abstract}

\newpage

\section{Introduction}

The ability to model and automatically detect discourse structure
is an important step toward understanding spontaneous dialogue.
While there is hardly consensus on exactly how discourse structure should
be described, some agreement exists that a useful
first level of analysis involves the identification of {\bf dialogue acts} (DAs).
A DA represents the meaning of an utterance at the
level of illocutionary force \cite{Austin:62}.
Thus, a DA is approximately the equivalent
of the speech act of \namecite{Searle:69},
the conversational game move of \namecite{Power:79},
or the adjacency pair part of
\namecite{Schegloff:68} and \namecite{Sacks:74}.

\begin{table}
\caption{\label{tab1}%
Fragment of a labeled conversation (from the Switchboard corpus).}
{\small
\begin{tabular}{cllll}
\hline
Speaker &\multicolumn{2}{c}{Dialogue Act}&\multicolumn{2}{c}{Utterance}\\ 
\hline
{\bf A} & \multicolumn{2}{l}{\DA{Yes-No-Question}} & \multicolumn{2}{l}{So do you go to college right now? }\\
{\bf A} & \multicolumn{2}{l}{\DA{Abandoned}} & \multicolumn{2}{l}{Are yo-, }\\
{\bf B} & \ ~~~~~~~~~~~& {\DA{Yes-Answer}}& ~~~ &{\em           Yeah,  }\\
{\bf B} & &\DA{Statement}&& {\em it's my last year {\footnotesize\em [laughter]}.} \\
{\bf A} & \multicolumn{2}{l}{\DA{Declarative-Question}} & \multicolumn{2}{l}{You're a, so you're a senior now.}\\
 {\bf B} & &\DA{Yes-Answer} & & {\em Yeah,}\\
{\bf B} & &\DA{Statement} && {\em I'm working on my projects trying to graduate {\footnotesize\em [laughter]}.}\\
{\bf A} & \multicolumn{2}{l}{\DA{Appreciation}} & \multicolumn{2}{l}{Oh, good for you.}\\
{\bf B} & &\DA{Backchannel} && {\em Yeah.} \\
{\bf A} & \multicolumn{2}{l}{\DA{Appreciation}} & \multicolumn{2}{l}{That's great, }\\
{\bf A} & \multicolumn{2}{l}{\DA{Yes-No-Question}} & \multicolumn{2}{l}{um, is, is N C University is that, uh, State,}\\
{\bf B} & &\DA{Statement} && {\em N C State.}\\
{\bf A} & \multicolumn{2}{l}{\DA{Signal-Non-Understanding}} & \multicolumn{2}{l}{What did you say? }\\
{\bf B} & &\DA{Statement} && {\em N C State. }\\
\end{tabular}
}
\end{table}

Table~\ref{tab1} shows a sample of the kind of discourse structure
in which we are interested.  Each utterance is assigned a unique
DA label (shown in column 2), drawn from a well-defined set
(shown in Table~\ref{tab2}).
Thus, DAs can be thought of as a tag set that classifies utterances 
according to a combination of pragmatic, semantic, and syntactic 
criteria. The computational community has usually
defined these DA categories so as to be relevant
to a particular application,
although efforts are under way to develop DA labeling systems that are
domain-independent, such as the Discourse Resource Initiative's
DAMSL architecture \cite{CoreAllen:97}.

\begin{table}
\caption{\label{tab2}%
The 42 dialogue act labels. DA frequencies are given as percentages of
the total number of utterances in the overall corpus.}
{\small
\begin{tabular}{llr@{}l}
\hline
{\bf Tag }&{\bf Example} & \multicolumn{2}{l}{\bf \%} \\
\hline
\DA{Statement}&{\em  Me,  I'm in the legal department.} & 36&\%\\
\DA{Backchannel/Acknowledge} & {\em Uh-huh.} & 19&\%\\
\DA{Opinion}&  {\em I think it's great}& 13&\%\\
\DA{Abandoned/Uninterpretable} &{\em So, -/}& 6&\% \\
\DA{Agreement/Accept}& {\em That's exactly it.} & 5&\%\\
\DA{Appreciation}&{\em I can imagine.}& 2&\%\\
\DA{Yes-No-Question}&{\em Do you have to have any special training?}& 2&\%\\
\DA{Non-verbal} & {\em $<$Laughter$>$,$<$Throat\_clearing$>$} & 2&\% \\
\DA{Yes answers} & {\em Yes.} & 1&\%\\
\DA{Conventional-closing}&{\em Well, it's been nice talking to you.}&1&\%\\
\DA{Wh-Question}&{\em What did you wear to work today?}& 1&\%\\
\DA{No answers}& {\em No.} & 1&\%\\
\DA{Response Acknowledgment}&{\em Oh, okay.}& 1&\% \\
\DA{Hedge}& {\footnotesize\em I don't know if I'm making any sense or not.} & 1&\% \\
\DA{Declarative Yes-No-Question} &{\em So you can afford to get a house?}& 1&\%\\
\DA{Other}& {\em  Well give me a break, you know. }& 1&\%\\
\DA{Backchannel-Question} & {\em Is that right?} & 1&\%\\
\DA{Quotation}& {\em You can't be pregnant and have cats} & &.5\%\\
\DA{Summarize/reformulate}& {\footnotesize\em Oh, you mean you switched schools for the kids.} & &.5\%\\
\DA{Affirmative non-yes answers}& {\em It is.} & &.4\%\\
\DA{Action-directive}          & {\em Why don't you go first}& &.4\%\\
\DA{Collaborative Completion}&{\em Who aren't contributing.}& &.4\%\\
\DA{Repeat-phrase}& {\em Oh, fajitas} & &.3\%\\
\DA{Open-Question}&{\em How about you?}& &.3\%\\
\DA{Rhetorical-Questions}&{\em Who would steal a newspaper?  }& &.2\%\\
\DA{Hold before answer/agreement}& {\em I'm drawing a blank.} & &.3\%\\
\DA{Reject}& {\em Well, no} & &.2\% \\
\DA{Negative non-no answers}& {\em  Uh,  not a whole lot.} & &.1\% \\
\DA{Signal-non-understanding}& {\em Excuse me?  } & &.1\%\\
\DA{Other answers}& {\em I don't know} & &.1\%\\
\DA{Conventional-opening}&{\em How are you?}& &.1\%\\
\DA{Or-Clause}&{\em or is it more of a company?} & &.1\%\\
\DA{Dispreferred answers}&{\em Well, not so much that.}& &.1\%\\
\DA{3rd-party-talk}    & {\em My goodness, Diane, get down from there.} & &.1\%\\
\DA{Offers, Options \& Commits}& {\em I'll have to check that out} & &.1\%\\
\DA{Self-talk}&{\em What's the word I'm looking for} & &.1\%\\
\DA{Downplayer}& {\em That's all right.} & &.1\%\\
\DA{Maybe/Accept-part}& {\em Something like that}& $<$&.1\%\\
\DA{Tag-Question}&{\em Right?} & $<$&.1\%\\
\DA{Declarative Wh-Question}&{\em You are what kind of buff?}& $<$&.1\%\\
\DA{Apology}& {\em I'm sorry.} & $<$&.1\% \\
\DA{Thanking}& {\em  Hey thanks a  lot} & $<$&.1\%\\
\end{tabular}
}
\end{table}

While not constituting dialogue understanding in any deep sense,
DA tagging seems clearly useful to a range of 
applications.  For example,
a meeting summarizer needs to keep track of who said what to whom,
and a conversational agent needs to know whether it was asked
a question or ordered to do something.
In related work DAs are used as a first processing step to infer dialogue
games~\cite{Carlson:83,LevinMoore:77,LevinEtAl:99}, a slightly higher level unit
that comprises a small number of DAs.
Interactional dominance~\cite{Linell:90} might be measured more
accurately using DA distributions than with simpler techniques,
and could serve as an indicator of the type or genre of discourse at hand.
In all these cases, DA labels would enrich the available input for
higher-level processing of the spoken words.
Another important role of DA information could be feedback to
lower-level processing.
For example, a speech recognizer could be constrained by expectations
of likely DAs in a given context, constraining the potential recognition
hypotheses so as to improve accuracy.

The goal of this article is twofold:
On the one hand, we aim to present a comprehensive framework
for modeling and automatic classification of DAs, founded on
well-known statistical methods. In doing so, we will pull together
previous approaches as well as new ideas.
For example, our model draws on the use of DA
$n$-grams and the hidden Markov models of conversation present in
earlier work, such as \namecite[1994]{NagataMorimoto:93}
\nocite{NagataMorimoto:94} and
\namecite{WoszczynaWaibel:94} (see Section~\ref{sec:prior}).
However, our framework generalizes earlier models,
giving us a clean probabilistic
approach for performing DA classification from unreliable words and
nonlexical evidence.
For the speech recognition task, our
framework provides a mathematically principled way to condition
the speech recognizer on conversation context through dialogue structure,
as well as on nonlexical information correlated with DA identity.
We will present methods in a domain-independent framework
that for the most part treats DA labels as an arbitrary formal tag set.
Throughout the presentation, we will highlight the simplifications
and assumptions made to achieve tractable models, and point out
how they might fall short of reality.

Second, we present results obtained with this approach on
a large, widely available corpus of spontaneous conversational speech.
These results, besides
validating the methods described, are of interest for several
reasons.  For example, unlike in most previous work on DA labeling,
the corpus is not task-oriented in nature, and the amount of data used
(198,000 utterances)
exceeds that in previous studies by at least an order of magnitude
(see Table~\ref{tabcomparison}).

To keep the presentation interesting and concrete, we
will alternate between the description of general methods and empirical results.
Section~\ref{sec:task} describes the task and our data in detail.
Section~\ref{sec:hmm} presents the probabilistic modeling framework;
a central component of this framework, the discourse grammar,
is further discussed in Section~\ref{sec:grammars}.
In Section~\ref{sec:detection} we describe experiments for
DA classification. Section~\ref{sec:asr} shows how DA models can
be used to benefit speech recognition.
Prior and related work is summarized in
Section~\ref{sec:prior}.
Further issues and open problems are addressed in Section~\ref{sec:discussion},
followed by concluding remarks in Section~\ref{sec:conclusion}.

\section{The Dialogue Act Labeling Task}
        \label{sec:task}

The domain we chose to model is the 
Switchboard corpus of human-human conversational telephone speech
\cite{SWBD} distributed by the Linguistic Data Consortium.
Each conversation involved two randomly selected strangers who had been charged 
with talking informally about one of several, self-selected general-interest
topics.
To train our statistical models on this corpus, we combined an
extensive
effort in human hand-coding of DAs for each utterance, together with a 
variety of automatic and semiautomatic tools. 
Our data consisted of a substantial portion of the Switchboard waveforms and
corresponding transcripts, totaling 1,155 conversations.

\cbstart
\subsection{Utterance Segmentation}
	\label{sec:segmentation}
\cbend

Before hand-labeling each utterance in the corpus with a DA,
we needed to choose an utterance segmentation, as
the raw Switchboard data is not segmented in a linguistically consistent
\cbstart
way.  To expedite the DA labeling task and remain consistent with 
other Switchboard-based research efforts,
we made use of a version of the corpus
that had been hand-segmented into sentence-level units prior to our own work
and independently of our DA labeling system \cite{SWBD-DF}.
We refer to the units of this segmentation as {\bf utterances}.
The relation between utterances and speaker turns is not one-to-one:
a single turn can contain multiple utterances, and 
utterances can span more than one turn (e.g., in the case of backchanneling
by the other speaker in mid-utterance).
Each utterance unit was identified with one DA, and was annotated with 
a single DA label.
The DA labeling system had special provisions for rare cases where
utterances seemed to combine aspects of several DA types.
\cbend

Automatic segmentation of spontaneous speech is an open research problem in
its own right \cite{MastEtAl:96,StoShr:icslp96}.
A rough idea of the difficulty of the segmentation problem on this corpus
and using the same definition of utterance units can be derived from 
a recent study \cite{ShribergEtAl:specom2000}.  In an automatic labeling
of word boundaries as either utterance or nonboundaries using a 
combination of lexical and prosodic cues, we obtained 96\% accuracy
based on correct word transcripts, and 78\% accuracy with automatically 
recognized words.
The fact that the segmentation and labeling tasks are
interdependent~\cite{WarnkeEtAl:97,FinkeEtAl:aaai98} 
further complicates the problem.

\cbstart
Based on these considerations, we decided not to confound
the DA classification task with the additional problems introduced by automatic
segmentation and assumed the utterance-level segmentations as given.
\cbend
An important consequence of this decision is that we can 
expect utterance length and acoustic properties at utterance boundaries to be
accurate, both of which turn out to be important features of DAs
\cite[see also Section~\ref{sec:prosody-features}]{ShribergEtAl:LS}.

\cbstart
\subsection{Tag Set}
	\label{sec:tagset}
\cbend

We chose to follow a recent standard for shallow discourse structure annotation,
the Dialogue Act Markup in Several Layers (DAMSL) tag set,
which was designed by the natural language processing community 
under the auspices of the Discourse Resource Initiative
\cite{CoreAllen:97}.  We began with the DAMSL markup system, but
\cbstart
modified it in several ways to make it more relevant to our corpus and task.
DAMSL aims to provide a domain-independent framework for dialogue annotation,
as reflected by the fact that our tag set can be mapped back to DAMSL
categories \cite{Jurafsky:97-damsl}.
However, our labeling effort also showed 
that content- and task-related distinctions
will always play an important role in effective DA labeling.
\cbend

The Switchboard domain itself is essentially ``task-free,'' thus giving
few external constraints on the definition of DA categories.  Our
primary purpose in adapting the tag set was to enable 
computational DA modeling for conversational
speech, with possible improvements to conversational speech
recognition.  Because of the lack of a specific task, we decided to label
categories that seemed both inherently interesting linguistically and
that could be identified reliably.  
Also, the focus on conversational speech
recognition led to a certain bias toward categories that were lexically
or syntactically distinct (recognition accuracy is traditionally measured
including all lexical elements in an utterance).

While the  modeling techniques described in this paper are formally
independent of the corpus and the choice of tag set, their success on
any particular task will of course crucially depend on these factors.
For different tasks not all the techniques used
in this study might prove useful and others could be of greater importance.
However, we believe that this study represents a fairly comprehensive
application of technology in this area and can serve as a point of
departure and reference for other work.

The resulting SWBD-DAMSL tag set was multidimensional;
approximately 50 basic tags (e.g., \DA{Question}, \DA{Statement}) could each be combined
with diacritics indicating orthogonal information, for example, about whether
or not the dialogue function of the utterance was related to Task-Management and
Communication-Management.
Approximately 220 of the many possible unique combinations of these
codes were used by the coders  \cite{Jurafsky:97-damsl}.  
To obtain a system with somewhat higher interlabeler agreement, as well as
enough data per class for statistical
modeling purposes, a less fine-grained tag set was devised.
This tag set distinguishes 42 mutually exclusive utterance types and was used
for the experiments reported here.
Table~\ref{tab2} shows the 42 categories with examples and relative
frequencies.%
\footnote{For the study focusing on prosodic modeling of DAs reported elsewhere
\cite{ShribergEtAl:LS}, the tag set was further reduced to six categories.}
While some of the original infrequent classes were collapsed, the resulting
DA type distribution is still highly skewed.
This occurs largely because there was no basis for subdividing the dominant
DA categories according to task-independent and reliable criteria.

The tag set incorporates both traditional sociolinguistic and
discourse-theoretic notions, such as
rhetorical relations and adjacency-pairs, as well as
some more form-based labels.  Furthermore, the tag set is structured so
as to allow labelers to annotate a Switchboard conversation from transcripts
alone (i.e., without listening) in about 30 minutes.
Without these constraints the DA labels might have included some
finer distinctions, but we felt that this drawback was balanced by the
ability to cover a large amount of data.%
\footnote{The effect of lacking acoustic information on labeling accuracy
was assessed by relabeling a subset of the data {\em with} listening,
and was found to be fairly small \cite{ShribergEtAl:LS}.
A conservative estimate based on the relabeling study is 
that for most DA types at most 2\% of the labels might have changed based
on listening.  The only DA types with higher uncertainty
were \DA{Backchannels} and \DA{Agreements}, which are easily confused
with each other
without acoustic cues; here the rate of change was no more than 10\%.
}

Labeling was carried out in a three-month period in 1997 by eight linguistics
graduate students at CU Boulder.  Interlabeler agreement for the
42-label tag set used here was 84\%, resulting in a Kappa statistic of 0.80.
The Kappa statistic measures agreement normalized for chance
\cite{Siegel:88}.
\cbstart
As argued in \namecite{Carletta:96}, Kappa values of 0.8 or higher are
desirable for detecting associations between several coded variables;
we were thus satisfied with the level of agreement achieved.
(Note that, even though only a single variable, DA type, was
coded for the present study, our goal is, among other things, to model
associations between several instances of that variable, e.g., between
adjacent DAs.)
\cbend

A total of 1,155 Switchboard conversations were labeled,
comprising 205,000 utterances and 1.4 million words.
The data was partitioned into a training set of 1,115 conversations
(1.4M words, 198K utterances), used for estimating the various components of
our model, and a test set of 19 conversations (29K words, 4K utterances).
Remaining conversations were set aside for future use
(e.g., as a test set uncompromised of tuning effects).

\subsection{Major Dialogue Act Types}

The more frequent DA types are briefly characterized below.
As discussed above, the focus of this paper is not on
the nature of DAs, but on the computational framework
for their recognition;
\cbstart
full details of the DA tag set and numerous
motivating examples can be found in a separate report
\cite{Jurafsky:97-damsl}.
\cbend

\paragraph{Statements and Opinions}
The most common types of utterances were
\DA{Statements}  and \DA{Opinions}.
This split distinguishes ``descriptive, narrative, or personal'' statements
(\DA{Statement})
from ``other-directed opinion statements'' (\DA{Opinion}).  The distinction was
designed to capture the different kinds of responses
we saw to opinions (which are often countered or disagreed with via
further opinions) and to statements (which more often elicit
continuers or backchannels):

\begin{center}
\begin{tabular}{ll}
\multicolumn{1}{c}{Dialogue Act}&\multicolumn{1}{c}{Example Utterance}\\ 
\hline
\DA{Statement} &     Well,  we have a cat, um, \\
\DA{Statement} &     He's probably, oh,  a  good two years old, \\
& big, old, fat and sassy tabby.\\
\DA{Statement} &  He's about five months old \\
\DA{Opinion} &     Well,  rabbits are darling.\\
\DA{Opinion} &     I think it would be kind of stressful.\\
\end{tabular}
\end{center}

\DA{Opinions} often include such hedges as {\em I think},
{\em I believe}, {\em it seems}, and {\em I mean}.
We combined the \DA{Statement} and 
\DA{Opinion} classes for other studies on dimensions
in which they did not differ \cite{ShribergEtAl:LS}.

\paragraph{Questions}

Questions were of several types.
The \DA{Yes-No-Question} label includes only utterances 
having both the pragmatic force of a yes-no-question {\em and}
the syntactic markings of a yes-no-question 
(i.e., subject-inversion or sentence-final tags).
\DA{Declarative-Questions}
are utterances that function pragmatically as questions but do not
have ``question form.''  By this we mean that
declarative questions normally have no {\em wh}-word as the argument of the verb 
(except in ``echo-question'' format), and have ``declarative'' word order 
in which the subject precedes the verb. See
\namecite{Weber:93} for a survey of declarative questions
and their various realizations.

\begin{center}
\begin{tabular}{ll}
\multicolumn{1}{c}{Dialogue Act}&\multicolumn{1}{c}{Example Utterance}\\ 
\hline
\DA{Yes-No-Question} & Do you have to have any special training? \\
\DA{Yes-No-Question }& But  that doesn't eliminate it, does it?\\
\DA{Yes-No-Question }& Uh,  I guess a year ago you're probably\\
& watching C N N a lot, right? \\
\DA{Declarative-Question }&   So you're taking a government course? \\
\DA{Wh-Question}& Well,  how old are you?\\
\end{tabular}
\end{center}

\paragraph{Backchannels}
A backchannel is a short utterance that plays discourse-structuring
roles, e.g., indicating that the speaker should go on talking.
These are usually referred to in the conversation analysis
literature as ``continuers'' and have been studied extensively
\cite{Jefferson:84,Schegloff:82,Yngve:70}.
We expect recognition of backchannels to be useful because of
their discourse-structuring role
(knowing that the hearer expects the speaker to go on talking
tells us something about the course of the narrative)
and because they seem to occur at certain kinds of syntactic boundaries;
detecting a backchannel may thus help in predicting utterance boundaries
and surrounding lexical material.

\begin{table}
\caption{Most common realizations of backchannels in Switchboard.}
\begin{tabular}{rlrlrl}\hline
{\bf Frequency}&{\bf Form}&
{\bf Frequency}&{\bf Form}&
{\bf Frequency}&{\bf Form}\\
\hline
38\% & uh-huh & 2\% &   yes & 1\% &   sure\\
34\% & yeah & 2\% &   okay & 1\% &   um\\
9\% &  right & 2\% &   oh yeah & 1\% &   huh-uh\\
3\% &   oh & 1\% &   huh & 1\% & uh\\
\end{tabular}
\label{tabback}
\end{table}

For an intuition about what backchannels look like, Table~\ref{tabback} shows
the most common realizations of the approximately 300 types
(35,827 tokens) of backchannel in our Switchboard subset.
The following table shows examples of backchannels in the context of a
Switchboard conversation:
\begin{center}
\begin{tabular}{cllll}
Speaker &\multicolumn{2}{c}{Dialogue Act}&\multicolumn{2}{c}{Utterance}\\ 
\hline
{\bf B}&\ ~~~~~~~~~~~&{\DA{Statement}} & ~~~ &{\em but, uh,  we're to the
point now where our}\\
& & & & {\em  financial income is enough that we can consider}   \\
& & & & {\em  putting some away --}   \\
{\bf A}&\multicolumn{2}{l}{\DA{Backchannel}} &\multicolumn{2}{l}{Uh-huh. /}  \\
{\bf B}&&\DA{Statement} &&{\em -- for college, / } \\
{\bf B}&& \DA{Statement}& &{\em so we are going to be starting a regular payroll} \\
& & & & {\em  deduction --}   \\
{\bf A}&\multicolumn{2}{l}{\DA{Backchannel}}&\multicolumn{2}{l}{Um. /}\\
{\bf B}&&\DA{Statement}&&{\em --- in the fall /  }\\
{\bf B}&&\DA{Statement}&&{\em and then the money that I will be making
this } \\
& & & & {\em summer we'll be putting away for the college}\\  
& & & & {\em fund.}\\  
{\bf A}&\multicolumn{2}{l}{\DA{Appreciation}}&\multicolumn{2}{l}{Um.
Sounds good.}    \\
\end{tabular}
\end{center}

\paragraph{Turn Exits and Abandoned Utterances}

Abandoned utterances are those that the speaker breaks off without
finishing, and are followed by a restart.  Turn exits resemble 
abandoned utterances in that they are often syntactically broken off,
but they are used mainly as a way of passing speakership to the other
speaker.  Turn exits tend to be single words, often {\em so} or {\em or}.
\begin{center}
\begin{tabular}{cllll}
Speaker &\multicolumn{2}{c}{Dialogue Act}&\multicolumn{2}{c}{Utterance}\\ 
\hline
{\bf A}&\multicolumn{2}{l}{\DA{Statement}} &\multicolumn{2}{l}{we're from,
uh,  I'm from  Ohio  /} \\
{\bf A}&\multicolumn{2}{l}{\DA{Statement}} &\multicolumn{2}{l}{and  my
wife's from Florida  /} \\
{\bf A}&\multicolumn{2}{l}{\DA{Turn-Exit}} &\multicolumn{2}{l}{so, -/} \\
{\bf B}& \ ~~~~&{\DA{Backchannel}} &\ ~~~~~&{\em Uh-huh. / } \\ 
\\
{\bf A}&\multicolumn{2}{l}{\DA{Hedge}} &\multicolumn{2}{l}{so, I don't know, /} \\
{\bf A}&\multicolumn{2}{l}{\DA{Abandoned}} &\multicolumn{2}{l}{it's $<$lipsmack$>$, - / }\\
{\bf A}&\multicolumn{2}{l}{\DA{Statement}} &\multicolumn{2}{l}{I'm glad
it's not the kind of problem I have to} \\
& & & \multicolumn{2}{l}{come up with an answer to because it's not --} \\ \hline
\end{tabular}
\end{center}

\paragraph{Answers and Agreements}

\DA{Yes-Answers} 
include {\em yes}, {\em yeah}, {\em yep}, {\em uh-huh}, and other
variations on {\em yes},
when they are acting as an answer to a \DA{Yes-No-Question} or 
\DA{Declarative-Question}.  Similarly, we also coded \DA{No-Answers}.
Detecting \DA{Answers} can help tell us that the previous
utterance was a \DA{Yes-No-Question}.  Answers are also semantically
significant since they are likely to contain new information.

\DA{Agreement/Accept}, \DA{Reject}, and \DA{Maybe/Accept-Part}
all mark the degree to which a speaker
accepts some previous proposal, plan, opinion, or statement.
The most common of these are the 
\DA{Agreement/Accepts}.
These are very often {\em yes} or {\em yeah},
so they look a lot like \DA{Answers}. But where answers follow
questions, agreements often follow opinions or proposals, so distinguishing
these can be important for the discourse.

\section{Hidden Markov Modeling of Dialogue}
        \label{sec:hmm}

We will now describe the mathematical and computational framework used in
our study.
Our goal is to perform DA classification and other tasks
using a probabilistic formulation,
giving us a principled approach for combining multiple knowledge sources
(using the laws of probability), as well as the ability to derive 
model parameters automatically from a corpus, using statistical inference
techniques.

Given all available evidence $E$ about a conversation,
the goal is to find the DA sequence $U$ that has
the highest posterior probability $P(U | E)$ given that evidence.
Applying Bayes' Rule we get
\begin{eqnarray}
 U^\ast & = & \argmax_{U} P(U | E)  \nonumber \\
        & = & \argmax_{U} { P(U) P(E | U) \over P(E) } \nonumber \\
        & = & \argmax_{U} P(U) P(E | U) \label{eq:bayes}
\end{eqnarray}
Here $P(U)$ represents the prior probability of a DA sequence,
and $P(E | U)$ is the likelihood of $U$ given the evidence.
The likelihood is usually much more straightforward to model than
the posterior itself. This has to do with the fact that our 
models are generative or causal in nature, i.e., they describe 
how the evidence is produced by the underlying DA sequence $U$.

Estimating $P(U)$ requires building a probabilistic {\bf discourse grammar},
i.e., a statistical model of DA sequences.  This can be done using familiar
techniques from language modeling for speech recognition, although
the sequenced objects in this case are DA labels rather than words;
discourse grammars will be discussed in detail in Section~\ref{sec:grammars}.

\subsection{Dialogue Act Likelihoods}

The computation of likelihoods $P(E | U)$
depends on the types of evidence used.  In our experiments we used
the following sources of evidence, either alone or in combination:
\begin{description}
\item[Transcribed words:]   
        The likelihoods used in Equation~\ref{eq:bayes} are $P(W | U)$,
        where $W$ refers to the true (hand-transcribed) words spoken
        in a conversation.
\item[Recognized words:]
        The evidence consists of recognizer acoustics $A$, and we
        seek to compute $P(A | U)$.  As described later, this 
        involves considering multiple alternative recognized 
        word sequences.
\item[Prosodic features:]
        Evidence is given by the acoustic features $F$ capturing various
        aspects of pitch, duration, energy, etc., of the speech signal;
        the associated likelihoods are $P(F | U)$.
\end{description}

\begin{table}
\caption{Summary of random variables used in dialogue modeling.
(Speaker labels are introduced in Section~\protect\ref{sec:grammars}.)}
\label{tab:rvar}
    \begin{tabular}{cl}
        \hline
        Symbol  & Meaning \\
        \hline
      $U$ & sequence of DA labels\\
      $E$ & evidence (complete speech signal)\\
      $F$ & prosodic evidence\\
      $A$ & acoustic evidence (spectral features used in ASR)\\
      $W$ & sequence of words\\
      $T$ & speakers labels\\
    \end{tabular}
\end{table}

For ease of reference, all random variables used here are summarized 
in Table~\ref{tab:rvar}.
The same variables are used with subscripts to refer to individual
utterances.
For example, $W_i$ is the word transcription of the $i$th utterance
within a conversation ({\em not} the $i$th word).

To make both the modeling and the search for the best DA sequence 
feasible, we further require that our likelihood models are
{\bf decomposable by utterance}.  
This means that the likelihood given a complete conversation can
be factored into likelihoods given the individual utterances.
We use $U_i$ for the $i$th DA label in the sequence $U$, i.e.,
$U = (U_1, \ldots, U_i, \ldots, U_n)$,
where $n$ is the number of utterances in a conversation.
In addition, we use $E_i$ for that portion of the evidence
that corresponds to the $i$th utterance, e.g., the words or the
prosody of the $i$th utterance.
Decomposability of the likelihood means that
\begin{equation}
        P(E | U) = P(E_1 | U_1) \cdot \ldots \cdot P(E_n | U_n)
\end{equation}

Applied separately to the three types of evidence $A_i$, $W_i$ and $F_i$ mentioned above,
it is clear that this assumption is not strictly true.
For example, speakers tend to reuse words found earlier
in the conversation \cite{FowlerHousum:87}
and an answer might actually be relevant to the question before it,
violating the independence of the $P(W_i | U_i)$.
Similarly, 
speakers adjust their pitch or volume over time, 
e.g., to the conversation partner or because of the structure 
of the discourse \cite{MennBoyce:82},
violating the independence of
the $P(F_i | U_i)$.
As in other areas of statistical modeling, we count on the fact
that these violations are
small compared to the properties actually
modeled, namely, the dependence of $E_i$ on $U_i$.

\subsection{Markov Modeling}

Returning to the prior distribution of DA sequences $P(U)$, it is convenient
to make certain independence assumptions here, too.
In particular, we assume that the prior distribution of $U$ is Markovian,
i.e., that each $U_i$ depends only on a fixed number $k$ of preceding
DA labels:
\begin{equation}
        P(U_i | U_1, \ldots, U_{i-1}) = P(U_i | U_{i-k}, \ldots, U_{i-1})
\end{equation}
($k$ is the order of the Markov process describing $U$).
The $n$-gram-based discourse grammars we used have this property.
As described later, $k = 1$ is a very good choice, i.e., conditioning
on the DA types more than one removed from the current one does not
improve the quality of the model by much, at least with the amount of
data available in our experiments.

The importance of the Markov assumption for the discourse grammar
is that we can now view the whole system of discourse grammar and
local utterance-based likelihoods as a $k$th-order hidden Markov model
(HMM) \cite{Rabiner:86}.
The HMM states correspond to DAs, observations correspond
to utterances, transition probabilities are given by the discourse grammar
(see Section~\ref{sec:grammars}),
and observation probabilities are given by the local likelihoods $P(E_i|U_i)$.

\begin{figure}
\lefteqn{
\begin{array}{ccccccc}
 & E_1 &                                        & E_i &                                        & E_n \\
 & \uparrow    &                                & \uparrow &                                   & \uparrow \\
\mbox{<start>} \longrightarrow
 & U_1 & \longrightarrow \cdots \longrightarrow & U_i & \longrightarrow \cdots \longrightarrow & U_n & \longrightarrow \mbox{<end>} \\
\end{array}
}

\caption{\label{fig:bayes1}%
The discourse HMM as Bayes network.}
\end{figure}

We can represent the dependency structure (as well as the implied conditional
independences) as a special case of Bayesian belief network \cite{Pearl:88}.
Figure~\ref{fig:bayes1} shows the variables in the resulting HMM with
directed edges representing conditional dependence.  To keep things simple,
a first-order HMM (bigram discourse grammar) is assumed.

\subsection{Dialogue Act Decoding}

The HMM representation allows us to use efficient dynamic programming
algorithms to compute relevant aspects of the model, such as
\begin{itemize}
\item
        the most probable DA sequence (the Viterbi algorithm)
\item
        the posterior probability of various DAs for a given utterance,
        after considering all the evidence (the forward-backward algorithm)
\end{itemize}

The Viterbi algorithm for HMMs \cite{Viterbi:67} finds the globally most
probable state
sequence.  When applied to a discourse model with locally decomposable
likelihoods and Markovian discourse grammar, it will therefore find 
precisely the DA sequence with the highest posterior probability:
\begin{equation}
        U^\ast = \argmax_{U} P(U | E)
\end{equation}
The combination of likelihood and prior modeling, HMMs, and Viterbi decoding
is fundamentally the same as the standard probabilistic approaches to
speech recognition \cite{Bahl:83} and tagging \cite{Church:88}.
It maximizes the probability of getting the {\em entire} DA sequence 
correct, but it does not necessarily find the DA sequence that 
has the most DA labels correct \cite{Dermatas:95}.
To minimize the total number of utterance labeling errors,
we need to maximize the 
probability of getting each DA label correct individually, i.e.,
we need to maximize $P(U_i | E)$ for each $i = 1, \ldots, n$.
We can compute the per-utterance posterior DA probabilities by
summing:
\begin{equation}
        P(u | E) = \sum_{U: U_i = u} P(U | E) \quad
\end{equation}
where the summation is over all sequences $U$ whose $i$th element
matches the label in question.
The summation is efficiently carried out by the forward-backward algorithm
for HMMs \cite{Baum:70}.%
\footnote{We note in passing that the Viterbi and Baum algorithms have 
equivalent formulations in the Bayes network framework \cite{Pearl:88}.
The HMM terminology was chosen here mainly for historical reasons.}

For 0th-order (unigram) discourse grammars, Viterbi decoding
and forward-backward decoding necessarily yield the same results.
However, for higher-order discourse grammars
we found that forward-backward decoding consistently gives slightly
(up to 1\% absolute) better accuracies, as expected.
Therefore, we used this method throughout.

The formulation presented here, as well as all our experiments,
uses the {\em entire conversation} as evidence for DA classification.
Obviously, this is possible only during offline
processing, when the full conversation is available.
Our paradigm thus follows historical practice in the Switchboard domain,
where the goal is typically the offline processing (e.g., automatic
transcription, speaker identification, indexing, archival) of entire previously
recorded conversations.
However, the HMM formulation used here also supports computing posterior DA
probabilities based on partial evidence, e.g., using only the utterances
preceding the current one, as would be required for online processing.

\section{Discourse Grammars}
        \label{sec:grammars}

The statistical discourse grammar models the prior probabilities $P(U)$ of
DA sequences.  In the case of conversations for which the identities 
of the speakers are known (as in Switchboard), the discourse grammar should
also model turn-taking behavior.  
A straightforward approach is to model sequences of pairs
$(U_i, T_i)$ where $U_i$ is the DA label and $T_i$ represents the 
speaker. We are not trying to model speaker idiosyncrasies, so
conversants are arbitrarily identified as {\bf A} or {\bf B}, and the model is 
made symmetric with respect to the choice of sides (e.g., by 
replicating the training sequences with sides switched).
Our discourse grammars thus had a vocabulary of $42 \times 2 = 84$ labels,
plus tags for the beginning and end of conversations.
For example, the second DA tag in Table~\ref{tab1} would be predicted 
by a trigram discourse grammar using the fact that the {\em same} speaker
previously uttered a \DA{Yes-No-Question}, which in turn was preceded by
the start-of-conversation.

\subsection{N-gram Discourse Models}

A computationally convenient type of discourse grammar is an $n$-gram model
based on DA tags, as it allows efficient decoding in the HMM framework.
We trained standard backoff $n$-gram models \cite{Katz:87}, using 
the frequency smoothing approach of \namecite{Witten:91}.
Models of various orders were compared by their perplexities,
i.e., the average number of choices the model predicts for each tag,
conditioned on the preceding tags.

\begin{table}
\caption{\label{tab:ppl}%
Perplexities of DAs with and without turn information.}
\begin{tabular}{lr@{}lr@{}lr@{}l}
\hline
Discourse Grammar       & \multicolumn{2}{c}{$P(U)$} &
				\multicolumn{2}{c}{$P(U,T)$} & 
					\multicolumn{2}{c}{$P(U|T)$} \\
\hline
None                    & 42&		& 84&		& 42& \\
Unigram                 & 11&.0		& 18&.5		& 9&.0 \\
Bigram                  & 7&.9		& 10&.4		& 5&.1 \\
Trigram                 & 7&.5		& 9&.8		& 4&.8  \\
\end{tabular}
\end{table}

Table~\ref{tab:ppl} shows perplexities for three types of models:
$P(U)$, the DAs alone; $P(U,T)$, the combined DA/speaker ID sequence;
and $P(U | T)$, the DAs conditioned on known speaker IDs
(appropriate for the Switchboard task).
As expected, we see an improvement (decreasing perplexities) for increasing
$n$-gram order.  However, the incremental gain of a trigram is small,
and higher-order models did not prove useful.
(This observation, initially based on perplexity, is confirmed by
the DA tagging experiments reported in Section~\ref{sec:detection}.)
Comparing $P(U)$ and $P(U |T)$, we see that speaker identity adds substantial
information, especially for higher-order models.

The relatively small improvements from higher-order models
could be a result of lack of training data, or of an inherent independence
of DAs from DAs further removed.
The near-optimality of the bigram discourse grammar
is plausible given conversation analysis accounts of
discourse structure in terms of adjacency pairs
\cite{Schegloff:68,Sacks:74}.
Inspection of bigram probabilities estimated from our data
revealed that conventional adjacency pairs receive high probabilities,
as expected.  For example,
30\% of \DA{Yes-No-Questions} are followed by \DA{Yes-Answers},
14\% by \DA{No-Answers} (confirming that the latter are dispreferred).
\DA{Commands} are followed by \DA{Agreements} in 23\% of the cases,
and \DA{Statements} elicit \DA{Backchannels} in 26\% of all
cases.

\subsection{Other Discourse Models}

We also investigated non-$n$-gram discourse models, based on various 
language modeling techniques known from speech recognition.
One motivation for alternative models is that $n$-grams enforce 
a one-dimensional representation on DA sequences, whereas we saw above
that the event space is really multidimensional
(DA label and speaker labels).  Another motivation is that 
$n$-grams fail to model long-distance dependencies, such as the fact
that speakers may tend to repeat certain DAs or patterns throughout
the conversation.

The first alternative approach was a standard cache model
\cite{KuhnDeMori:90}, which boosts the probabilities of previously observed
unigrams and bigrams, on the theory that tokens tend to repeat themselves
over longer distances.
However, this does not seem to be true for DA sequences in our corpus,
as the cache model showed no improvement over the standard $n$-gram.
This result is somewhat surprising since unigram dialogue grammars are able to detect
speaker gender with 63\% accuracy (over a 50\% baseline) on
Switchboard~\cite{Ries:99b}, indicating
that there {\em are} global variables in the DA distribution that could
potentially be exploited by a cache dialogue grammar.
Clearly, dialogue grammar adaptation needs further research.

Second, we built a discourse grammar that incorporated
constraints on DA sequences in a nonhierarchical way, using 
maximum entropy (ME) estimation \cite{Berger:96}.
The choice of features was informed by similar ones commonly used in
statistical language models, as well our general intuitions about potentially
information-bearing elements in the discourse context.
Thus, the model was designed so that the current DA label was constrained
by features such as unigram statistics, the previous DA and the DA once
removed, DAs occurring within a window in the past, and whether the previous
utterance was by the same speaker.
We found, however, that an ME model using $n$-gram constraints performed
only slightly better than a corresponding backoff $n$-gram.

Additional constraints such as DA triggers, distance-1 bigrams, separate
encoding of speaker change and bigrams to the last DA on the same/other
channel did not improve relative  to the trigram model.
The ME model thus confirms the adequacy of the backoff $n$-gram approach,
and leads us to conclude that DA sequences, at least in the Switchboard domain,
are mostly characterized by local
interactions, and thus modeled well by low-order $n$-gram statistics for this
task.
For more structured tasks this situation might be different.
However, we have found no further exploitable structure.

\section{Dialogue Act Classification}
        \label{sec:detection}

We now describe in more detail how the knowledge sources
of words and prosody are modeled, and what automatic DA labeling results
were obtained using each of the knowledge sources in turn.  Finally,
we present results for a combination of all knowledge sources.
DA labeling accuracy results should be compared to a baseline
(chance) accuracy of 35\%, the relative frequency of the most frequent DA type 
(\DA{Statement}) in our test set.\footnote{The frequency
of \DA{Statements} across all labeled data was slightly different,
cf.~Table~\ref{tab2}.}

\subsection{Dialogue Act Classification Using Words}
        \label{sec:from-words}

DA classification using words is based on the observation that different
DAs use distinctive word strings.
It is known that certain cue words and phrases \cite{HirschbergLitman:93}
can serve as explicit indicators of discourse structure.
Similarly, we find distinctive correlations between certain phrases
and DA types.
For example, 92.4\% of the {\em uh-huh}'s occur in \DA{Backchannels}, and
88.4\% of the trigrams ``$<$start$>$ {\em do you}'' occur in \DA{Yes-No-Questions}.
To leverage this information source, without hand-coding knowledge
about which words are indicative of which DAs, we will use statistical
language models that model the full word sequences associated with
each DA type.

\subsubsection{Classification from True Words}
        \label{sec:from-true-words}

Assuming that the true (hand-transcribed) words of utterances are given 
as evidence, we can compute word-based likelihoods $P(W | U)$
in a straightforward way, by building a statistical language model
for each of the 42 DAs.
All DAs of a particular type found in the training corpus 
were pooled, and a DA-specific trigram model was estimated using standard 
techniques ({\citebrackets Katz-backoff \cite{Katz:87} with Witten-Bell discounting
\cite{Witten:91}}).

\subsubsection{Classification from Recognized Words}

For fully automatic DA classification, the above approach is only a 
partial solution, since we are not yet able to recognize words in
spontaneous speech with perfect accuracy.
A standard approach is to use the 1-best hypothesis from the speech
recognizer in place of the true word transcripts.  While conceptually
simple and convenient, this method will not make optimal use of all
the information in the recognizer, which in fact maintains multiple 
hypotheses as well as their relative plausibilities.

A more thorough use of recognized speech can be derived as follows.
The classification framework is modified such that the recognizer's acoustic
information (spectral features) $A$ appear as the evidence.
We compute $P(A | U)$ by decomposing it into an acoustic likelihood
$P(A | W)$ and a word-based likelihood $P(W | U)$, and summing over all
word sequences:
\begin{eqnarray}
        \label{eq:from-nbest}
        P(A | U) & = & \sum_{W} P(A | W, U) P(W | U) \nonumber \\
            & = & \sum_{W} P(A | W) P(W | U) 
\end{eqnarray}
The second line is justified under the assumption that the recognizer
acoustics (typically, cepstral coefficients) are invariant to 
DA type once the words are fixed.
Note that this is another approximation in our modeling.
For example, different DAs with common words may be realized by 
different word pronunciations.
Figure~\ref{fig:bayes-with-words} shows the Bayes network resulting from
modeling recognizer acoustics through word hypotheses under this 
independence assumption; note the added $W_i$
variables (that have to be summed over) in comparison
to Figure~\ref{fig:bayes1}.

\begin{figure}
\lefteqn{
\begin{array}{ccccccc}
 & A_1 &                                        & A_i &                                        & A_n \\
 & \uparrow    &                                & \uparrow &                                   & \uparrow \\
 & W_1 &                                        & W_i &                                        & W_n \\
 & \uparrow    &                                & \uparrow &                                   & \uparrow \\
\mbox{<start>} \longrightarrow
 & U_1 & \longrightarrow \cdots \longrightarrow & U_i & \longrightarrow \cdots \longrightarrow & U_n & \longrightarrow \mbox{<end>} \\
\end{array}
}

\caption{\label{fig:bayes-with-words}%
Modified Bayes network including word hypotheses and recognizer
acoustics.}
\end{figure}

The acoustic likelihoods $P(A | W)$ correspond to the acoustic scores
the recognizer outputs for every hypothesized word sequence $W$.
The summation over all $W$ must be approximated;
in our experiments we summed over the (up to) 2,500 best hypotheses generated
by the recognizer for each utterance.  Care must be taken to scale 
the recognizer acoustic scores properly, i.e., to exponentiate the recognizer 
acoustic scores by $1/\lambda$, where $\lambda$ is the language model weight
of the recognizer.\footnote{\label{fn:score-scaling}
In a standard recognizer the total log score of a hypothesis $W_i$
is computed as
\[
        \log P(A_i | W_i) + \lambda \log P(W_i) - \mu |{W_i}| \quad ,
\]
where $|{W_i}|$ is the number of words in the hypothesis,
and both $\lambda$ and $\mu$ are parameters optimized to minimize the word
error rate.
The word insertion penalty $\mu$ represents a correction
to the language model that allows balancing insertion and deletion errors.
The language model weight $\lambda$ compensates for acoustic scores
variances that are effectively too large due to severe independence assumptions
in the recognizer acoustic model.
According to this rationale, it is more appropriate
to {\em divide} all score components by $\lambda$. Thus, in all 
our experiments, we computed a summand in Equation~\ref{eq:from-nbest} whose
logarithm was 
\[
        {1 \over \lambda} \log P(A_i | W_i) + \log P(W_i | U_i) -
                        {\mu \over \lambda } |{W_i}| \quad .
\]
We found this approach to give better results than the 
standard multiplication of $\log P(W)$ by $\lambda$.
Note that for selecting the best hypothesis in a recognizer
only the relative magnitudes of the score weights matter;
however, for the summation in Equation~\ref{eq:from-nbest} the absolute values
become important.
The parameter values for $\lambda$ and $\mu$ were those used 
by the standard recognizer; they were not specifically optimized for 
the DA classification task.}

\subsubsection{Results}

Table~\ref{tab:word-results} shows DA classification accuracies
obtained by combining the word- and recognizer-based likelihoods 
with the $n$-gram discourse grammars described earlier.
The best accuracy obtained from transcribed words, 71\%, is encouraging
given a comparable human performance of 84\% (the interlabeler agreement,
see Section~\ref{sec:tagset}).
We observe about a 21\% relative increase in classification error
when using recognizer words;
this is remarkably small considering that the speech recognizer used had a
word error rate of 41\% on the test set.

\begin{table}
\caption{\label{tab:word-results}%
DA classification accuracies (in \%) from transcribed and recognized words
(chance $=$ 35\%).}
\begin{tabular}{lccc}
\hline
Discourse Grammar &     True            & Recognized & Relative Error Increase\\
\hline
None                    & 54.3          & 42.8   & 25.2\%       \\
Unigram                 & 68.2          & 61.8   & 20.1\%       \\
Bigram                  & 70.6          & 64.3   & 21.4\%       \\
Trigram                 & 71.0          & 64.8   & 21.4\%       \\
\hline
\end{tabular}
\end{table}

We also compared the $n$-best DA classification approach to the more 
straightforward 1-best approach.
In this experiment, only the single best recognizer
hypothesis is used, effectively treating it as the true word string.
The 1-best method increased classification error by about 7\% relative
to the $n$-best algorithm (61.5\% accuracy with a bigram discourse grammar).

\subsection{Dialogue Act Classification Using Prosody}
        \label{sec:from-prosody}

We also investigated prosodic information, i.e., information independent
of the words as well as the standard recognizer acoustics.
Prosody is important for DA recognition for two reasons.
First, as we saw earlier, word-based classification suffers
from recognition errors.
Second, some utterances are inherently ambiguous based on words alone.
For example, some \DA{Yes-No-Questions} have word sequences identical to
those of \DA{Statements}, but can often be distinguished by their final F0 rise.

\cbstart
A detailed study aimed at automatic prosodic classification of DAs
in the Switchboard domain is available in a companion paper
\cite{ShribergEtAl:LS}.
Here we investigate the interaction of prosodic models with the
dialogue grammar and the word-based DA models discussed above.
We also touch briefly on alternative machine learning models for 
prosodic features.
\cbend

\subsubsection{Prosodic Features}        \label{sec:prosody-features}
        
Prosodic DA classification was based on a large set of features
computed automatically from the waveform, without reference to word or
phone information.  The features can be broadly grouped as referring
to duration (e.g., utterance duration, with and without pauses),
pauses (e.g., total and mean of nonspeech regions exceeding 100~ms),
pitch (e.g., mean and range of F0 over utterance, slope of F0
regression line), energy (e.g., mean and range of RMS energy, same for
signal-to-noise ratio [SNR]), speaking rate
({\citebrackets based on the ``enrate'' measure
of \namecite{Morgan-enrate:97}}), and gender (of both speaker and
listener).
In the case of utterance duration, the measure correlates
both with length in words and with overall speaking rate.  The gender
feature that classified speakers as either male or female was used to
test for potential inadequacies in F0 normalizations.
Where appropriate, we included both raw features and values normalized
by utterance and/or conversation.  We also included features that are
the output of the pitch accent and boundary tone event detector of
\namecite{Taylor:2000} (e.g., the number of pitch accents in the
utterance).
\cbstart
A complete description of prosodic features and an
analysis of their usage in our models can be found
in \namecite{ShribergEtAl:LS}.
\cbend

\subsubsection{Prosodic Decision Trees}

For our prosodic classifiers, we used CART-style decision trees
\cite{Breiman:84}. Decision trees allow combination of discrete and
continuous features, and can be inspected to help in understanding 
the role of different features and feature combinations.

\begin{figure}
\strut\psfig{file=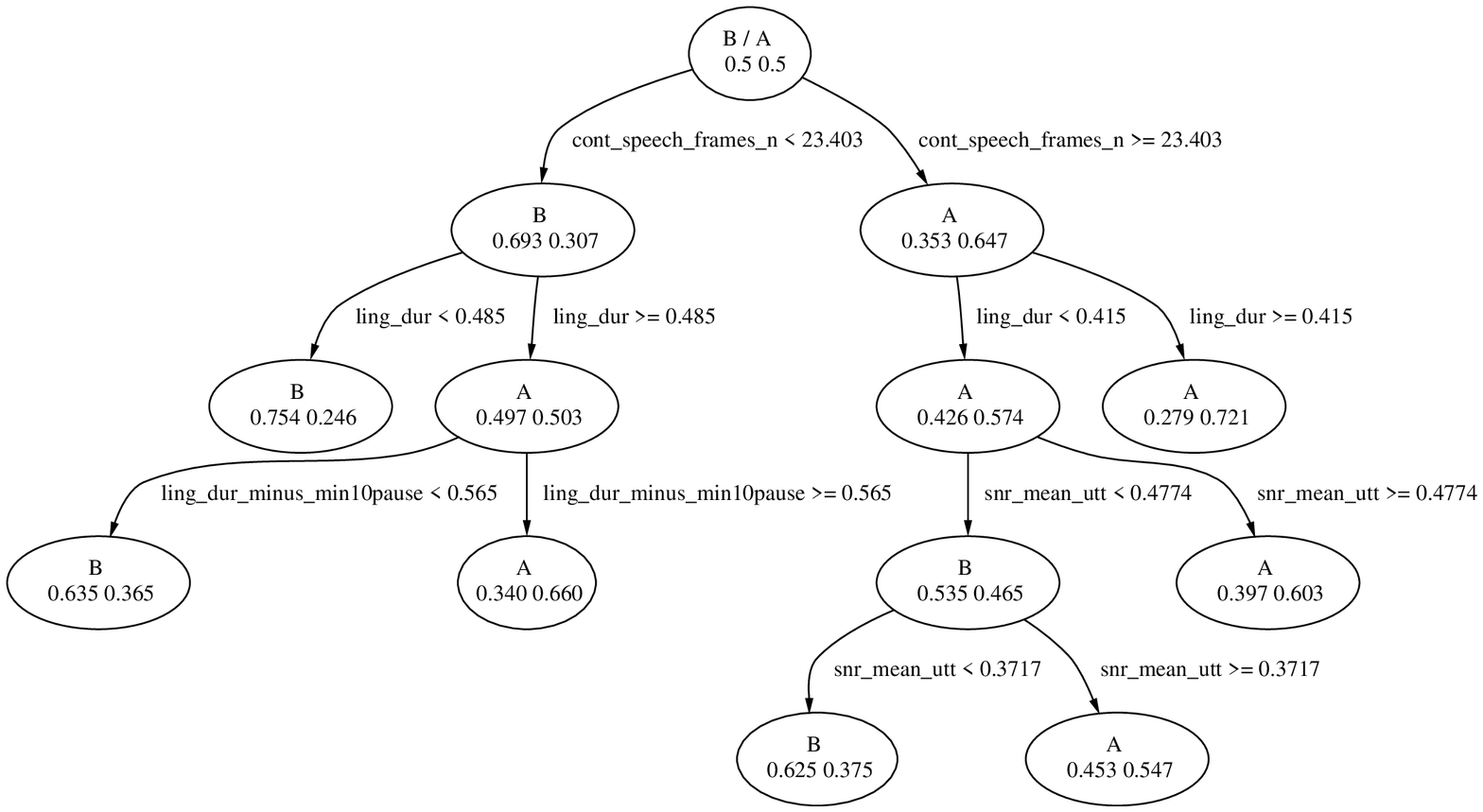,angle=270,width=\textwidth}
\caption{\label{fig:tree-145}%
Decision tree for the classification of \DA{Backchannels} (B) and \DA{Agreements} (A).
Each node is labeled with the majority class for that node, as well as
the posterior probabilities of the two classes.
The following features are queried in the tree:
number of frames in continuous ($>1$ s)
speech regions ({\em cont\_speech\_frames}),
total utterance duration ({\em ling\_dir}),
utterance duration excluding pauses $>100$ ms ({\em ling\_dur\_minus\_min10pause}),
and mean signal-to-noise ratio ({\em snr\_mean\_utt}).
}
\end{figure}

To illustrate one area in which prosody could aid our classification
task, we applied trees to DA classifications known to be ambiguous from
words alone.  One frequent example in our corpus was the distinction
between \DA{Backchannels} and \DA{Agreements}
(see Table~\ref{tab2}), which share terms
such as {\em right} and {\em yeah}. As shown in Figure~\ref{fig:tree-145},
a prosodic tree trained on this task revealed that agreements have consistently
longer durations and greater energy (as reflected by the SNR measure)
than do backchannels.

The HMM framework requires that we compute prosodic likelihoods
of the form $P(F_i | U_i)$ for each utterance $U_i$ and associated
prosodic feature values $F_i$.
We have the apparent difficulty that decision trees (as well as
other classifiers, such as neural networks) give
estimates for the posterior probabilities, $P(U_i | F_i)$.
The problem can be overcome by applying Bayes' Rule locally:
\begin{equation}
        P(F_i| U_i) = P(F_i) { P(U_i | F_i) \over P(U_i) } 
                        \propto { P(U_i|F_i) \over P(U_i) }
\end{equation}
Note that $P(F_i)$ does not depend on $U_i$ and can be 
treated as a constant for the purpose of DA classification.
A quantity proportional to the required likelihood can therefore
be obtained either by dividing the posterior tree probability
by the prior $P(U_i)$,\footnote{
\namecite{Bourlard:93} use this approach to integrate neural network
phonetic models in a speech recognizer.}
or by training the tree on a uniform prior distribution
of DA types.  We chose the second approach, downsampling our training 
data to equate DA proportions.  This also counteracts a common problem
with tree classifiers trained on very skewed distributions of target classes,
i.e., that low-frequency classes are not modeled in sufficient detail
because the majority class dominates the tree-growing objective function.

\subsubsection{Results with Decision Trees}
        \label{sec:prosody-results}

As a preliminary experiment to test the integration of prosody with
other knowledge sources, we trained a
single tree to discriminate among the five most frequent
DA types (\DA{Statement}, \DA{Backchannel}, \DA{Opinion}, \DA{Abandoned},
and \DA{Agreement},
totaling 79\% of the data) and an Other category comprising all
remaining DA types.
The decision tree was trained on a downsampled training subset containing 
equal proportions of these six DA classes.
The tree achieved a classification accuracy of 45.4\% on an independent test
set with the same uniform six-class distribution.
The chance accuracy on this set is 16.6\%, so the tree clearly extracts 
useful information from the prosodic features.

We then used the decision tree posteriors as scaled DA likelihoods in the
dialogue model HMM, combining it with various $n$-gram dialogue grammars
for testing on our full standard test set.
For the purpose of model integration, the likelihoods of the Other
class were assigned to all DA types comprised by that class.
As shown in Table~\ref{tab:detect-prosody}, the tree with dialogue grammar
performs significantly better than chance on the raw DA distribution,
although not as well as the word-based methods
(cf.\ Table~\ref{tab:word-results}).

\begin{table}
\caption{DA classification using prosodic decision trees (chance $=$ 35\%).
\label{tab:detect-prosody}}
\begin{tabular}{lc}
\hline
Discourse Grammar       &       Accuracy (\%)   \\
\hline
None                    &       38.9    \\
Unigram                 &       48.3    \\
Bigram                  &       49.7    \\
\hline
\end{tabular}
\end{table}

\subsubsection{Neural Network Classifiers}

Although we chose to use decision trees as prosodic classifiers for 
their relative ease of inspection,
we might have used any suitable probabilistic classifier,
i.e., any model that estimates the posterior probabilities of DAs
given the prosodic features. We conducted preliminary experiments
to assess how neural networks compare to decision trees for the type
of data studied here.  Neural networks are worth investigating since
they offer potential advantages over decision trees.  They
can learn decision surfaces that lie at an angle to the axes of the
input feature space, unlike standard CART trees, which always split
continuous features on one dimension at a time.  The response function
of neural networks is continuous (smooth) at the decision boundaries,
allowing them to avoid hard decisions and the complete fragmentation
of data associated with decision tree questions.  

Most important, however, related work~\cite{Ries:99} indicated that similarly
structured
networks are superior classifiers if the input features are words and are
therefore a plug-in replacement for the language model classifiers described in
this paper.
Neural networks are therefore a good candidate for a jointly optimized
classifier of prosodic and word-level information since one can show that
they are a generalization of the integration approach used here.

\begin{table}
\caption{Performance of various prosodic neural network classifiers
on an equal-priors, six-class DA set (chance $=$ 16.6\%).
\label{tab:nn-results}}
\begin{tabular}{lc}
\hline
Network Architecture            &       Accuracy (\%)   \\
\hline
Decision tree   &       45.4  \\
\\
No hidden layer, linear output function & 44.6        \\
No hidden layer, softmax output function & 46.0       \\
40-unit hidden layer, softmax output function & 46.0  \\
\hline
\end{tabular}
\end{table}

We tested various neural network models on the same six-class downsampled
data as used for decision tree training, using a variety of 
network architectures and output layer functions.
The results are summarized in Table~\ref{tab:nn-results},
along with the baseline result obtained with the decision tree model.
Based on these experiments, a softmax network \cite{Bridle:90}
without hidden units resulted in only a slight improvement
over the decision tree.  A network with hidden units did
not afford any additional advantage, even after we optimized the number
of hidden units, indicating that complex combinations of
features (as far as the network could learn them) do not 
predict DAs better than linear combinations of input features.
While we believe alternative classifier architectures should be 
investigated further as prosodic models, the results so far seem to confirm
our choice of decision trees as a model class that gives close to optimal
performance for this task.

\subsubsection{Intonation Event Likelihoods}

An alternative way to compute prosodically based DA likelihoods 
uses pitch accents and boundary phrases \cite{Taylor:97}.
The approach relies on the intuition that different utterance types
are characterized by different intonational ``tunes'' \cite{Kowtko:96},
and has been successfully applied to the classification of move types in
\cbstart
the DCIEM Map Task corpus \cite{WrightTaylor:97}.
\cbend
The system detects sequences of
distinctive pitch patterns by training one continuous-density HMM for each
DA type.  Unfortunately, the event classification accuracy on the Switchboard
corpus was considerably poorer than in the Map Task domain, and DA recognition
results when coupled with a discourse grammar were substantially worse
than with decision trees.  The approach could prove valuable in the future,
however, if the intonation event detector can be made more robust to
corpora like ours.

\subsection{Using Multiple Knowledge Sources}
        \label{sec:from-all}

As mentioned earlier, we expect improved performance from combining 
word and prosodic information.
Combining these knowledge sources requires estimating a combined 
likelihood $P(A_i , F_i | U_i)$ for each utterance.
The simplest approach is to assume that the two types of acoustic 
observations (recognizer acoustics and prosodic features) are
approximately conditionally independent once $U_i$ is given:
\begin{eqnarray}
        P(A_i , W_i, F_i | U_i) & = & P(A_i, W_i | U_i) P( F_i | A_i, W_i, U_i)
                                                        \nonumber \\
                & \approx & P(A_i, W_i | U_i) P( F_i | U_i)
                                                \label{eq:multiple-ks}
\end{eqnarray}
Since the recognizer acoustics are modeled by way of their dependence
on words, it is particularly important to avoid using prosodic 
features that are directly correlated with word identities, or 
features that are also modeled by the discourse grammars, such
as utterance position relative to turn changes.
Figure~\ref{fig:bayes-with-multiple} depicts the Bayes network incorporating
evidence from both word recognition and prosodic features.

\begin{figure}
\lefteqn{
\begin{array}{ccccccc}
 & A_1 &                                        & A_i &                                        & A_n \\
 & \uparrow    &                                & \uparrow &                                   & \uparrow \\
 & W_1 &                                        & W_i &                                        & W_n \\
 & \uparrow    &                                & \uparrow &                                   & \uparrow \\
\mbox{<start>} \longrightarrow
 & U_1 & \longrightarrow \cdots \longrightarrow & U_i & \longrightarrow \cdots \longrightarrow & U_n & \longrightarrow \mbox{<end>} \\
 & \downarrow    &                              & \downarrow &                                 & \downarrow \\
 & F_1 &                                        & F_i &                                        & F_n \\
\end{array}
}

\caption{\label{fig:bayes-with-multiple}%
Bayes network for discourse HMM incorporating both word recognition and
prosodic features.}
\end{figure}

One important respect in which the independence assumption is violated is in
the modeling of utterance length.
While utterance length itself is not a prosodic feature, it is an important
feature to condition on when examining prosodic characteristics of 
utterances, and is thus best included in the decision tree.
Utterance length is captured directly by the tree using various
duration measures, while the DA-specific
LMs encode the average number of words per utterance indirectly through
$n$-gram parameters, but still accurately enough to violate 
independence in a significant way \cite{FinkeEtAl:aaai98}.
As discussed in Section~\ref{sec:discussion}, this problem is best
addressed by joint lexical-prosodic models.

We need to allow for the fact that the models combined in
Equation~\ref{eq:multiple-ks} give estimates of differing qualities.
Therefore, we introduce an exponential weight $\alpha$ on $P(F_i | U_i)$
that controls the contribution of the prosodic likelihood to the overall
likelihood.
Finally, a second exponential weight $\beta$ on the combined likelihood 
controls its dynamic range relative to the discourse grammar scores,
partially compensating for any correlation between the two likelihoods.
The revised combined likelihood estimate thus becomes
\begin{equation}
        P(A_i , W_i, F_i | U_i)  \approx 
                 \{ P(A_i, W_i | U_i) P( F_i | U_i)^{\alpha} \}^{\beta}
\end{equation}
In our experiments, the parameters $\alpha$ and $\beta$ were optimized 
using twofold jackknifing.  The test data was split roughly in half
(without speaker overlap),
each half was used to separately optimize the parameters, and the best values
were then tested on the respective other half. The reported results are from
the aggregate outcome on the two test set halves.

\subsubsection{Results}

In this experiment we combined
the acoustic $n$-best likelihoods based on recognized words
with the Top-5 tree classifier mentioned in Section~\ref{sec:prosody-results}.
Results are summarized in Table~\ref{tab:detect-combo}.

\begin{table}
\caption{\label{tab:detect-combo}%
Combined utterance classification accuracies (chance $=$ 35\%).
The first two columns correspond to Tables~\protect\ref{tab:detect-prosody}
and~\protect\ref{tab:word-results}, respectively.}
\begin{tabular}{lccc}
\hline
Discourse Grammar               & \multicolumn{3}{c}{Accuracy (\%)} \\
				\cline{2-4}
				& Prosody & Recognizer  & Combined \\
\hline
None                    &       38.9    & 42.8          & 56.5  \\
Unigram                 &       48.3    & 61.8          & 62.4  \\
Bigram                  &       49.7    & 64.3          & 65.0  \\
\end{tabular}
\end{table}

As shown, the combined classifier presents a slight improvement over the 
recognizer-based classifier.  The experiment without discourse grammar
indicates that the combined evidence is considerably stronger than
either knowledge source alone, yet this improvement seems to
be made largely redundant by the use of priors and the discourse grammar.
For example, by definition \DA{Declarative-Questions} are not
marked by syntax (e.g., by subject-auxiliary inversion) and are thus
confusable with \DA{Statements} and \DA{Opinions}.
While prosody is expected to help disambiguate
these cases, the ambiguity can also be removed by examining the
context of the utterance, e.g., by noticing that the following utterance
is a \DA{Yes-Answer} or \DA{No-Answer}.

\subsubsection{Focused Classifications}
        \label{sec:focussed}

To gain a better understanding of the potential for prosodic DA
classification independent of the effects of discourse grammar and 
the skewed DA distribution in Switchboard, we examined 
several binary DA classification tasks.  The choice of 
tasks was motivated by an analysis of confusions committed by
a purely word-based DA detector, which tends to
mistake \DA{Questions} for \DA{Statements}, and \DA{Backchannels} for \DA{Agreements}
(and vice versa).
We tested a prosodic classifier, a word-based classifier (with both
transcribed and recognized words), and a combined classifier
on these two tasks, downsampling the DA distribution to equate the
class sizes in each case.  Chance performance in all experiments
is therefore 50\%. Results are summarized in
Table~\ref{tab:integration-subtasks}.

\begin{table}
\caption{\label{tab:integration-subtasks}%
Accuracy (in \%) for individual and combined models for two subtasks,
using uniform priors (chance $=$ 50\%).}
\begin{tabular}{rcc}
\hline
\multicolumn{1}{l}{Classification Task} & True Words  & Recognized Words \\
\multicolumn{1}{r}{Knowledge Source}&     &   \\  
\hline
\multicolumn{1}{l}{\DA{Questions}/\DA{Statements}} & & \\
\quad prosody only              & 76.0   & 76.0  \\
\quad words only            & 85.9 & 75.4  \\
\quad words+prosody        & 87.6  & 79.8 \\  
\hline
\multicolumn{1}{l}{\DA{Agreements}/\DA{Backchannels}} & & \\
\quad prosody only          & 72.9  & 72.9 \\
\quad words only            & 81.0   & 78.2 \\
\quad words+prosody        & 84.7  & 81.7 \\
\end{tabular}
\end{table}

As shown, the combined classifier was consistently more accurate than
the classifier using words alone. Although the gain in accuracy was
not statistically significant for the small recognizer test set because of
a lack of power, replication for a larger hand-transcribed test set
showed the gain to
be highly significant for both subtasks by a Sign test,
\cbstart
$p<.001$ and $p<.0001$ (one-tailed), respectively.
\cbend
Across these, as well as additional subtasks,
the relative advantage of adding prosody was larger for recognized than for
true words, suggesting that prosody is particularly helpful when word
information is not perfect.

\section{Speech Recognition}
        \label{sec:asr}

We now consider ways to use DA modeling to enhance automatic speech
recognition (ASR).  The intuition behind this approach is that discourse
context
constrains the choice of DAs for a given utterance, and the DA type in turn
constrains the choice of words.  The latter can then
be leveraged for more accurate speech recognition.

\subsection{Integrating DA Modeling and ASR}
Constraints on the word sequences hypothesized by a recognizer are 
expressed probabilistically in the recognizer language model (LM).
It provides the prior distribution $P(W_i)$ for finding the a posteriori
most probable hypothesized words for an utterance, given the
acoustic evidence $A_i$ \cite{Bahl:83}:%
\footnote{Note the similarity of Equations~\ref{eq:asr} and~\ref{eq:bayes}.
They are identical except for the fact that we are now operating at the
level of an individual utterance, the evidence is given by the acoustics,
and the targets are word hypotheses instead of DA hypotheses.}
\begin{eqnarray}
 W_i^\ast & = & \argmax_{W_i} P(W_i | A_i)  \nonumber \\
        & = & \argmax_{W_i} { P(W_i) P(A_i | W_i) \over P(A_i) } \nonumber \\
        & = & \argmax_{W_i} P(W_i) P(A_i | W_i) \label{eq:asr}
\end{eqnarray}
The likelihoods $P(A_i | W_i)$ are estimated by the recognizer's
acoustic model. 
In a standard recognizer the language model $P(W_i)$ is the same for
all utterances; the idea here is to obtain better-quality
LMs by conditioning on the DA type $U_i$, since presumably the word 
distributions differ depending on DA type.
\begin{eqnarray}
 W_i^\ast & = & \argmax_{W_i} P(W_i | A_i, U_i)  \nonumber \\
        & = & \argmax_{W_i} { P(W_i | U_i) P(A_i | W_i, U_i)
                                \over P(A_i | U_i) } \nonumber \\
        & \approx & \argmax_{W_i} P(W_i | U_i) P(A_i | W_i) \label{eq:asr2}
\end{eqnarray}
As before in the DA classification model, we tacitly assume that the words $W_i$
depend only on the DA of the current utterance, and also that
the acoustics are independent of the DA type if the words are fixed.
The DA-conditioned language models $P(W_i | U_i)$ are readily trained 
from DA-specific training data, much like we did for DA classification from
words.\footnote{
In Equation~\ref{eq:asr2} and elsewhere in this section we gloss over
the issue of proper weighting of model probabilities, which is
extremely important in practice.  The approach explained in detail
in footnote~\ref{fn:score-scaling} applies here as well.}

The problem with applying Equation~\ref{eq:asr2}, of course, is that the DA
type $U_i$ is generally
not known (except maybe in applications where the user interface can be
engineered to allow only one kind of DA for a given utterance).
Therefore, we need to infer the likely DA types for each utterance,
using available evidence $E$ from the entire conversation.
This leads to the following formulation:
\begin{eqnarray}
 W_i^\ast & = & \argmax_{W_i} P(W_i | A_i, E) \nonumber \\
        & = & \argmax_{W_i} \sum_{U_i} P(W_i | A_i, U_i, E) P(U_i | E) \nonumber\\
        & \approx & \argmax_{W_i} \sum_{U_i} P(W_i | A_i, U_i) P(U_i | E)
                                                \label{eq:posterior-mix}
\end{eqnarray}
The last step in Equation~\ref{eq:posterior-mix} is justified because,
as shown in Figures~\ref{fig:bayes1} and~\ref{fig:bayes-with-multiple},
the evidence $E$ (acoustics, prosody, words) pertaining
to utterances other than $i$
can affect the current utterance only through its DA type $U_i$.

We call this the {\bf mixture-of-posteriors} approach, because it amounts
to a mixture of the posterior distributions obtained from DA-specific speech
recognizers (Equation~\ref{eq:asr2}), using the DA posteriors as weights.
This approach is quite expensive, however, as it requires multiple full
recognizer or rescoring passes of the input, one for each DA type.

A more efficient, though mathematically less accurate, solution can 
be obtained by combining guesses about the correct DA types directly at the
level of the LM.  
We estimate the distribution of likely DA types for a given utterance
using the entire conversation $E$ as evidence, and then use a
sentence-level mixture \cite{Iyer:94}
of DA-specific LMs in a {\em single} recognizer run.  In other words,
we replace $P(W_i | U_i)$ in Equation~\ref{eq:asr2} with
\[
        \sum_{U_i} P(W_i | U_i) P(U_i | E)      \quad ,
\]
a weighted mixture of all DA-specific LMs.
We call this the {\bf mixture-of-LMs} approach.
In practice, we would first estimate DA posteriors for each utterance,
using the forward-backward algorithm and the models described in
Section~\ref{sec:detection}, and then rerecognize the conversation or
rescore the recognizer output, using the new posterior-weighted mixture LM.
Fortunately, as shown in the next section, the mixture-of-LMs approach
seems to give results that are
almost identical (and as good) the mixture-of-posteriors approach.

\subsection{Computational Structure of Mixture Modeling}

It is instructive to compare the expanded scoring formulae
for the two DA mixture modeling approaches for ASR.
The mixture-of-posteriors approach yields
\begin{equation}
        P(W_i | A_i, E) = \sum_{U_i} { P(W_i|U_i) P(A_i|W_i) \over
                                        P(A_i | U_i) } P(U_i | E) \quad ,
\end{equation}
whereas the mixture-of-LMs approach gives
\begin{equation}
        P(W_i | A_i, E) \approx \left ( \sum_{U_i} P(W_i|U_i) P(U_i|E) \right )
                                        { P(A_i | W_i) \over P(A_i) } \quad .
\end{equation}
We see that the second equation reduces to the first under the crude 
approximation $P(A_i | U_i) \approx P(A_i)$.  In practice, the denominators
are computed by summing the numerators over a finite number of word hypotheses
$W_i$, so this difference translates into normalizing either after or
before summing over DAs.
When the normalization takes place as the final step it can by omitted for
score maximization purposes;  this shows why the mixture-of-LMs approach
is less computationally expensive.

\subsection{Experiments and Results}

We tested both the mixture-of-posteriors and the mixture-of-LMs approaches
on our Switchboard test set of 19 conversations.  Instead of decoding
the data from scratch using the modified models, we manipulated $n$-best
lists consisting of up to 2,500 best hypotheses for each utterance.
This approach is also convenient since both approaches require access to
the full word string for hypothesis scoring; the overall model is
no longer Markovian, and is therefore inconvenient to use in the
first decoding stage, or even in lattice rescoring.

The baseline for our experiments was obtained with a standard backoff trigram
language model estimated from all available training data.
The DA-specific language models were trained on word transcripts of all the
training utterances of a given type, and then smoothed further by interpolating
them with the baseline LM.  Each DA-specific LM used its own interpolation
weight, obtained by minimizing the perplexity of the interpolated model
on held-out DA-specific training data.  Note that this smoothing step is
helpful when using the DA-specific LMs for {\em word} recognition, but not for
DA classification, since it renders the DA-specific LMs less discriminative.%
\footnote{Indeed, during our DA classification experiments, we had observed
that smoothed DA-specific LMs yield lower classification accuracy.}

Table~\ref{tab:we} summarizes both the word error rates achieved with
the various models and the perplexities of the corresponding LMs used
in the rescoring (note that perplexity is not meaningful in the
mixture-of-posteriors approach).
For comparison, we also included two additional models:  the ``1-best LM''
refers to always using the DA-specific LM corresponding to the most probable
DA type for each utterance.  It is thus an approximation to both mixture
approaches where only the top DA is considered.
Second, we included an ``oracle LM,'' i.e., always using the LM that
corresponds to the {\em hand-labeled} DA for each utterance.  The purpose
of this experiment was to give us an upper bound on the effectiveness
of the mixture approaches, by assuming perfect DA recognition.

\begin{table}
\caption{Switchboard word recognition error rates and LM perplexities.}
\label{tab:we}
\begin{tabular}{lcc}
\hline
Model           & WER (\%)      & Perplexity    \\
\hline
Baseline        & 41.2                  & 76.8          \\
1-best LM       & 41.0                  & 69.3          \\
Mixture-of-posteriors   & 41.0          & n/a           \\
Mixture-of-LMs  & 40.9                  & 66.9          \\
Oracle LM       & 40.3                  & 66.8          \\
\end{tabular}
\end{table}

It was somewhat disappointing that the word error rate (WER) improvement in
the oracle experiment
was small (2.2\% relative), even though statistically highly significant
\cbstart
($p < .0001$, one-tailed, according to a Sign test on matched utterance pairs).
\cbend
The WER reduction achieved with the mixture-of-LMs approach did not achieve
statistical significance ($0.25 > p > 0.20$).  The 1-best DA and the
two mixture models also did not differ significantly on this test set.
In interpreting these results one must realize, however, that 
WER results depend on a complex combination of factors, most notably 
interaction between language models and the acoustic models. 
Since the experiments only varied the language models used in
rescoring, it is also informative to compare the quality of these
models as reflected by perplexity.  On this measure, we see a
substantial 13\% (relative) reduction,
which is achieved by both the oracle and the mixture-of-LMs.
The perplexity reduction for the 1-best LM is only 9.8\%,
showing the advantage of the mixture approach.

\begin{table}
\caption{Word error reductions through DA oracle, by DA type.}
\label{tab:oracle}
\begin{tabular}{lcr@{}lr@{}l}
\hline
Dialogue Act & Baseline WER & Oracle & \ WER & \multicolumn{2}{l}{WER Reduction} \\
\hline
\DA{No-Answer} & 29.4 & 11&.8 & -17&.6 \\
\DA{Backchannel} & 25.9 & 18&.6 & -7&.3 \\
\DA{Backchannel-Question} & 15.2 & 9&.1 & -6&.1 \\
\DA{Abandoned/Uninterpretable} & 48.9 & 45&.2 & -3&.7 \\
\DA{Wh-Question} & 38.4 & 34&.9 & -3&.5 \\
\DA{Yes-No-Question} & 55.5 & 52&.3 & -3&.2 \\
\DA{Statement} & 42.0 & 41&.5 & -0&.5 \\
\DA{Opinion} & 40.8 & 40&.4 & -0&.4 \\
\end{tabular}
\end{table}

\begin{figure}
\strut\psfig{figure=pie2.epsi,height=2in,clip=}
\caption{Relative contributions to test set word counts by DA type.}
\label{fig:pie}
\end{figure}

To better understand the lack of a more substantial reduction in word error,
we analyzed the effect of the DA-conditioned rescoring on the
individual DAs, i.e., grouping the test utterances by their true DA types.
Table~\ref{tab:oracle} shows the WER improvements for a few DA types,
ordered by the magnitude of improvement achieved.  As shown,
all frequent DA types saw improvement, but the highest wins were observed 
for typically short DAs, such as \DA{Answers} and \DA{Backchannels}.  This is
to be expected, as such DAs tend to be syntactically and lexically highly
constrained.
Furthermore, the distribution of number of words across DA types is very uneven
(Figure~\ref{fig:pie}).
\DA{Statements} and \DA{Opinions}, the DA types dominating in both frequency and number
of words (83\% of total), see no more than 0.5\% absolute improvement,
thus explaining the small overall improvement.
In hindsight, this is also not surprising, since the bulk of the
training data for the baseline LM consists of these DAs, allowing only 
little improvement in the DA-specific LMs.
A more detailed analysis of the effect of
DA modeling on speech recognition errors can be found elsewhere
\cite{VanEss:98}.

In summary, our experiments confirmed that DA modeling 
can improve word recognition accuracy quite substantially
\cbstart
in principle, at least for certain DA types,
\cbend
but that the skewed distribution of DAs (especially in terms
of number of words per type) limits the usefulness of the approach on
the Switchboard corpus.  
The benefits of DA modeling might therefore be
more pronounced on corpora with more even DA distribution, as is typically
the case for task-oriented dialogues.  
Task-oriented dialogues might also feature specific subtypes of general
DA categories that might be constrained by discourse.
Prior research on 
task-oriented dialogues summarized in the next section, however,
has also found only small reductions in WER (on the order of 1\%).
\cbstart
This suggests that even in task-oriented domains more research is needed to
realize the potential of DA modeling for ASR.
\cbend

\section{Prior and Related Work}
        \label{sec:prior}

As indicated in the introduction, our work builds on
a number of previous efforts in computational discourse modeling and
automatic discourse processing, most of which occurred over the
last half-decade.  
It is generally not possible to directly compare
quantitative results because of vast differences in methodology, tag set,
type and amount of training data, and, principally, assumptions made about what
information is available for ``free''
(e.g., hand-transcribed versus automatically recognized words,
or segmented versus unsegmented utterances).
Thus, we will focus on the conceptual aspects of previous research efforts,
and while we do offer a summary of previous quantitative results,
these should be interpreted as informative datapoints only, and 
not as fair comparisons between algorithms.

Previous research on DA modeling has generally focused on task-oriented dialogue,
with three tasks in particular garnering much of the research effort.
The Map Task corpus \cite{AndersonEtAl:91,BardEtAl:95} consists of
conversations between
two speakers with slightly different maps of an imaginary territory.
Their task is to help one speaker reproduce a route drawn
only on the other speaker's map,  all without being able to see each
other's maps.   Of the DA modeling algorithms described below,
\namecite{TaylorEtAl:LS98} and \namecite{Wright:98}
were based on Map Task.
The VERBMOBIL corpus consists of  two-party scheduling
dialogues.    A number of the DA modeling algorithms
described below were developed for VERBMOBIL, including those of
\namecite{MastEtAl:96},
\namecite{WarnkeEtAl:97},
\namecite{Reithinger:96}, 
\namecite{Reithinger:97}, and
\namecite{Samuel:98}.
The ATR Conference corpus is  a subset of a larger ATR Dialogue database
consisting of simulated dialogues
between a secretary and a questioner at international conferences.
Researchers using this corpus include \namecite{Nagata:92},
\namecite[1994]{NagataMorimoto:93}, \nocite{NagataMorimoto:94}
and \namecite{KitaEtAl:96}. Table~\ref{tab:tagsets}
shows the most commonly used versions of the tag sets from those three tasks.

\begin{table}
\caption{\label{tab:tagsets}%
Dialogue act tag sets used in three other extensively studied corpora.}
\hrule
\begin{description}
\item[VERBMOBIL.] These 18 high-level DAs used in VERBMOBIL-1 are 
abstracted over a total of 43 more specific DAs;
most experiments on VERBMOBIL DAs use the set of 18 rather than 43.
Examples are from \protect\namecite{JekatEtAl:95}.\\
{\small
\begin{tabular}{ll}
{\bf Tag }&{\bf Example}\\
\hline
\DA{Thank}&{\em  Thanks} \\
\DA{Greet} & {\em Hello Dan} \\
\DA{Introduce}&  {\em It's me again}\\
\DA{Bye} &{\em Alright bye}\\ 
\DA{Request-Comment} &{\em How does that look?}\\
\DA{Suggest} &{\em from thirteenth through seventeenth June}\\
\DA{Reject} &{\em No Friday I'm booked all day }\\
\DA{Accept} &{\em Saturday sounds fine, }\\
\DA{Request-Suggest} &{\em What is a good day of the week for you?}\\
\DA{Init} &{\em I wanted to make an appointment with you }\\
\DA{Give\_Reason} &{\em Because I have meetings all afternoon }\\
\DA{Feedback} &{\em Okay }\\
\DA{Deliberate} &{\em Let me check my calendar here}\\
\DA{Confirm} &{\em Okay, that would be wonderful}\\
\DA{Clarify} &{\em Okay, do you mean Tuesday the 23rd?}\\
\DA{Digress} &{\em [we could meet for lunch] and eat lots of ice cream}\\
\DA{Motivate} &{\em We should go to visit our subsidiary in Munich}\\
\DA{Garbage} &{\em Oops, I- }\\
\end{tabular}
}
\item[Map Task.] The 12 DAs or ``move types'' used in Map Task.
Examples are from \protect\namecite{TaylorEtAl:LS98}.\\
{\small
\begin{tabular}{ll}
{\bf Tag }&{\bf Example}\\
\hline
\DA{Instruct}&{\em  Go round, ehm horizontally underneath diamond mine} \\
\DA{Explain} & {\em I don't have a ravine} \\
\DA{Align}&  {\em Okay?}\\
\DA{Check} &{\em So going down to Indian Country?}\\ 
\DA{Query-yn} &{\em Have you got the graveyard written down?}\\
\DA{Query-w} &{\em In where?}\\
\DA{Acknowledge} &{\em Okay }\\
\DA{Clarify} &{\em $\{$you want to go... diagonally$\}$ Diagonally down}\\
\DA{Reply-y} &{\em I do.}\\
\DA{Reply-n} &{\em No, I don't}\\
\DA{Reply-w} &{\em $\{$And across to?$\}$ The pyramid.}\\
\DA{Ready} &{\em Okay }\\
\end{tabular}
}
\item[ATR.]
The 9 DAs (``illocutionary force types'') used in the ATR Dialogue
Database task;
some later models used an extended set of 15 DAs.
Examples are from the English translations given by
\protect\namecite{Nagata:92}.\\
{\small
\begin{tabular}{ll}
{\bf Tag }&{\bf Example}\\
\hline
\DA{Phatic}&{\em  Hello} \\
\DA{Expressive} & {\em Thank you} \\
\DA{Response}&  {\em That's right}\\
\DA{Promise} &{\em I will send you a registration form}\\ 
\DA{Request} &{\em Please go to Kitaooji station by subway}\\
\DA{Inform} &{\em We are not giving any discount this time}\\
\DA{Questionif} &{\em Do you have the announcement of the conference? }\\
\DA{Questionref} &{\em What should I do?}\\
\DA{Questionconf} &{\em You have already transferred the registration fee, right?}\\
\end{tabular}
}
\end{description}
\end{table}

As discussed earlier, these
domains differ from the Switchboard corpus in being
task-oriented.  Their tag sets are also generally smaller,
but some of the same problems of balance occur.
For example, in the Map Task domain, 33\%
of the words occur in 1 of the 12 DAs (\DA{instruct}).
Table~\ref{tabcomparison} shows the approximate size of the corpora, the
tag set, and tag estimation accuracy rates for various recent models of
DA prediction.  
\cbstart
The results summarized in the table also illustrate the differences in
inherent difficulty of the tasks.
\cbend
For example, the task of \namecite{WarnkeEtAl:97} was to simultaneously segment
and tag DAs, whereas the other results rely on a prior manual
segmentation. Similarly, the task in \namecite{Wright:98} and in our
study was to determine DA types from speech input, whereas work by others
is based on hand-transcribed textual input.

\begin{table}[p]
\caption{\label{tabcomparison}%
Data on recent DA tagging experiments.  The number of DA tokens reflects
training set size; accuracy refers to automatic tagging
correctness. The error rates should not be compared, since the tasks were
quite different.  The comment field indicates special difficulties due to
the type of input data.}
{\small
\begin{tabular}{lr@{}lr@{ }lr@{}lc}
\hline
{\bf Source} &\multicolumn{2}{l}{\bf Number of DA Tokens} & \multicolumn{2}{c}{\bf Number of DA Types/Tag Set} &\multicolumn{2}{c}{\bf Accuracy} & {\bf Comments} \\
\hline
\namecite{WoszczynaWaibel:94}   & 150--250&(?) & 6 & &  74&.1\% & \\
\namecite{NagataMorimoto:94}    & 2,450& &  15 &/ ATR & 39&.7\% & \\
\namecite{Reithinger:96}        & 6,494& &  18 &/ VERBMOBIL  &$\approx$ 40&\% & \\
\namecite{MastEtAl:96}          & 6,494& &  18 &/ VERBMOBIL & 59&.7\% & \\
\namecite{WarnkeEtAl:97}        & 6,494& & 18 &/ VERBMOBIL &53&.4\% & unsegmented \\
\namecite{Reithinger:97}        & 2,701&  & 18 &/ VERBMOBIL  &74&.7\% & \\
\namecite{Chu-Carroll:98}       & 915&   & 15 &           & 49&.71\% & \\
\namecite{Wright:98}            & 3,276& & 12 &/ Map Task & 64&\% & from speech\\
\namecite{TaylorEtAl:LS98}      & 9,272&   & 12 &/ Map Task  &47&\% & \\
\namecite{Samuel:98}		& 2,701&   & 18 &/ VERBMOBIL  &75&.12\% & \\
\namecite{FukadaEtAl:98}	& 3,584&	& 26 &/ C-Star (Japanese) & 81&.2\% & \\
\namecite{FukadaEtAl:98}	& 1,902& & 26 &/ C-Star (English)	& 56&.9\% & \\
Present study                   & 198,000&  & 42 &/ SWBD-DAMSL &65&\% & from speech\\
\end{tabular}
}
\end{table}

The use of $n$-grams to model the probabilities
of DA sequences, or to predict upcoming DAs online, 
has been proposed by many authors.
It seems to have been first employed  by \namecite{Nagata:92},
and in follow-up papers by \namecite[1994]{NagataMorimoto:93}
\nocite{NagataMorimoto:94} on the ATR Dialogue database.
The model predicted upcoming DAs
by using bigrams and trigrams conditioned on preceding DAs,
trained on a corpus of 2,722 DAs.    
Many others subsequently relied on and enhanced this $n$-grams-of-DAs approach,
often by applying standard techniques from statistical language modeling.
\namecite{Reithinger:96}, for example, used deleted interpolation
to smooth the dialogue $n$-grams.
\namecite{Chu-Carroll:98} uses knowledge of 
subdialogue structure to selectively skip previous
DAs in choosing conditioning for DA prediction.

\namecite[1994]{NagataMorimoto:93} \nocite{NagataMorimoto:94}
may also have been the first to
use word $n$-grams as a miniature grammar for DAs,
to be used in improving speech recognition.
The idea caught on very quickly:
\namecite{SuhmWaibel:94},
\namecite{MastEtAl:96},
\namecite{WarnkeEtAl:97},
\namecite{Reithinger:97}, and
\namecite{TaylorEtAl:LS98}
all use variants of backoff, interpolated, or class $n$-gram language models
to estimate DA likelihoods.
Any kind of sufficiently powerful, trainable language model could
perform this function, of course, and indeed
\namecite{Alexandersson:97} propose using automatically learned
stochastic context-free grammars.
Jurafsky, Shriberg, Fox, and Curl \shortcite{JurShrEtAl:98}
show that the grammar of some
DAs, such as appreciations, can be captured by finite-state automata over
part-of-speech tags.

$N$-gram models are {\em likelihood} models for DAs, i.e., they compute
the conditional
probabilities of the word sequence given the DA type.
Word-based {\em posterior probability} estimators are also possible, although
less common.  \namecite{MastEtAl:96} propose the use of semantic
classification trees, a kind of decision tree conditioned on word patterns
as features.
Finally, \namecite{Ries:99} shows that neural networks using
only unigram features can be superior to higher-order $n$-gram DA models.
\namecite{WarnkeEtAl:99} and \namecite{OhlerEtAl:99} use related
discriminative training algorithms for language models.

\namecite{WoszczynaWaibel:94}  and
\namecite{SuhmWaibel:94},
followed by \namecite{Chu-Carroll:98},
seem to have been the first to note that
such a combination of word and dialogue $n$-grams
could be viewed as a dialogue HMM with word strings as the observations.
(Indeed, with the exception of \namecite{Samuel:98},
all models listed in
Table~\ref{tabcomparison} rely on some version of this HMM metaphor.)
Some researchers explicitly used HMM induction techniques
to infer dialogue grammars.
\namecite{WoszczynaWaibel:94}, for example, trained an ergodic HMM 
using expectation-maximization to model speech act sequencing.
\namecite{KitaEtAl:96} made one of the few
attempts at {\em unsupervised} discovery of dialogue
structure, where a finite-state grammar induction algorithm is used to
find the topology of the dialogue grammar.

Computational approaches to prosodic modeling of DAs have aimed to
automatically extract various prosodic parameters---such as duration,
pitch, and energy patterns---from the speech signal
({\citebrackets \namecite{YoshimuraEtAl:96}; \namecite{Taylor:97};
\namecite{Kompe:97}, among others}).
Some approaches model
F0 patterns with techniques such as vector quantization and Gaussian
classifiers to help disambiguate utterance types.
An extensive comparison of the prosodic DA modeling literature with
our work can be found in \namecite{ShribergEtAl:LS}.

DA modeling has mostly been geared toward automatic DA classification,
and much less work has been done on applying DA models 
to automatic speech recognition.
\namecite{NagataMorimoto:94} suggest conditioning word language models
on DAs to lower perplexity.
\namecite{SuhmWaibel:94}  and
\namecite{EckertEtAl:96} each
condition a recognizer LM on left-to-right
DA predictions and are able to show reductions in word error rate of 1\%
on task-oriented corpora.
Most similar to our own work, but still in a task-oriented domain,
the work by \namecite{TaylorEtAl:LS98} combines
DA likelihoods from prosodic models with those from 1-best recognition output
to condition the recognizer LM, again achieving an absolute reduction
in word error rate of 1\%, similarly disappointing as the 0.3\% improvement
in our experiments.

Related computational tasks beyond DA classification and speech recognition
have received even less attention to date.
We already mentioned \namecite{WarnkeEtAl:97} and
\namecite{FinkeEtAl:aaai98}, who both showed that utterance segmentation and
classification can be integrated into a single search process.
\namecite{FukadaEtAl:98} investigate augmenting DA tagging with more detailed
semantic ``concept'' tags, as a preliminary step toward an
interlingua-based dialogue translation system.
\namecite{LevinEtAl:99} couple DA classification with dialogue game classification;
dialogue games are units above the DA level, i.e., short DA sequences such as
question-answer pairs.

All the work mentioned so far uses statistical models of various kinds.
As we have shown here, such models offer some fundamental advantages, such
as modularity and composability (e.g., of discourse grammars with DA models)
and the ability to deal with noisy input (e.g., from a speech recognizer)
in a principled way.  However, many other classifier architectures are 
applicable to the tasks discussed, in particular to
DA classification.  A nonprobabilistic approach for DA labeling proposed by
\namecite{Samuel:98} is transformation-based learning \cite{Brill:93}.
Finally it should be noted that there are other tasks with a mathematical
structure similar to that of DA tagging, such as shallow parsing for
natural language processing \cite{Munk:99} 
and DNA classification tasks~\cite{OhlerEtAl:99},
from which further techniques could be borrowed.

How does the approach presented here differ from these various
earlier models, particularly those based on HMMs?  
Apart from corpus and tag set differences,
our approach differs primarily in that it generalizes
the simple HMM approach to cope with new kinds of
problems, based on the Bayes network representations depicted in
Figures~\ref{fig:bayes-with-words} and
\ref{fig:bayes-with-multiple}.  For the DA classification task, our framework
allows us to do classification given unreliable words (by marginalizing over
the possible word strings corresponding to the acoustic input) and
given nonlexical (e.g., prosodic) evidence.
For the speech recognition task, the generalized model
gives a clean probabilistic framework for conditioning word probabilities
on the conversation context via the underlying DA structure.
Unlike previous models that did not address speech recognition or
relied only on an intuitive 1-best approximation, our model allows
computation of the optimum word sequence by effectively summing over all
possible DA sequences as well as all recognition hypotheses throughout
the conversation, using evidence from both past and future.

\section{Discussion and Issues for Future Research}
        \label{sec:discussion}

Our approach to dialogue modeling 
has two major components: statistical dialogue grammars modeling 
the sequencing of DAs, and DA likelihood models
expressing the local cues (both lexical and prosodic) for DAs.
We made a number of significant simplifications to arrive at
a computationally and statistically tractable formulation.
In this formulation, DAs serve as the hinges that join the
various model components, but also decouple these components through
statistical independence assumptions.  Conditional on the DAs,
the observations across utterances are assumed to be independent,
and evidence of different kinds from the same utterance (e.g., lexical and
prosodic) is assumed to be independent.  Finally, DA types themselves
are assumed independent beyond a short span (corresponding to the order of
the dialogue $n$-gram). 
Further research within this framework can be characterized by
which of these simplifications are addressed.

Dialogue grammars for conversational speech need to
be made more aware of the temporal properties of utterances.
For example, we are currently not modeling the fact that utterances
by the conversants may actually overlap (e.g., backchannels interrupting an
ongoing utterance).
In addition, we should model more of the nonlocal aspects of discourse
structure, despite our negative results so far.  For example, a context-free 
discourse grammar could potentially account for the nested structures
proposed in \namecite{Grosz:86}.\footnote{
The inadequacy of $n$-gram models for nested discourse structures is pointed out
by \namecite{Chu-Carroll:98}, although the suggested solution is 
a modified $n$-gram approach.}

The standard $n$-gram models for DA discrimination with lexical cues
are probably suboptimal for this task, simply because they are trained
in the maximum likelihood framework, without explicitly optimizing 
discrimination between DA types.
This may be overcome by using discriminative training procedures
\cite{WarnkeEtAl:99,OhlerEtAl:99}.
Training neural networks directly with posterior probability~\cite{Ries:99}
seems to be a more principled approach and it also offers much easier
integration with other knowledge sources. Prosodic features, for example,
can simply be added to the lexical features, allowing the model to
capture dependencies and redundancies across knowledge sources.
Keyword-based techniques from the field of 
message classification should also be applicable here \cite{Rose:91}.
Eventually, it is desirable to integrate dialogue grammar, lexical and
prosodic cues into a single model, e.g., one that predicts the 
next DA based on DA history and all the local evidence.

The study of automatically extracted prosodic features for DA modeling
is likewise only in its infancy.   Our preliminary experiments 
with neural networks have shown that small gains are obtainable with
improved statistical modeling techniques.  However, we believe that 
more progress can be made by improving the underlying features themselves,
in terms of both better understanding of how speakers use them,
and ways to reliably extract them from data.

Regarding the data itself, we saw that the distribution of DAs in our 
corpus limits the benefit of DA modeling for lower-level processing,
in particular
speech recognition.  The reason for the skewed distribution was in the
nature of the task (or lack thereof) in Switchboard.  It remains to 
be seen if more fine-grained DA distinctions can be made reliably 
in this corpus.  However, it should be noted that the DA definitions 
are really arbitrary as far as tasks other than DA labeling are 
concerned.  This suggests using unsupervised, self-organizing learning
schemes that choose their own DA definitions in the process of optimizing
the primary task, whatever it may be.  Hand-labeled DA categories may
still serve an important role in initializing such an algorithm.

We believe that dialogue-related tasks have much to benefit from
corpus-driven, automatic learning techniques.  To enable such research,
we need fairly large, standardized corpora that allow
comparisons over time and across approaches.  Despite its
shortcomings, the Switchboard domain could serve this purpose.

\section{Conclusions}
        \label{sec:conclusion}

We have developed an integrated probabilistic approach to dialogue act
modeling for conversational speech, and tested it on a large speech
corpus. The approach combines models for lexical and prosodic
realizations of DAs, as well as a statistical discourse grammar.  All
components of the model are automatically trained, and are thus
applicable to other domains for which labeled data is available.
Classification accuracies achieved so far are highly encouraging, relative
to the inherent difficulty of the task as measured by human labeler
performance.  We investigated several modeling alternatives for the
components of the model (backoff $n$-grams and maximum entropy models for
discourse grammars, decision trees and neural networks for prosodic
classification) and found performance largely independent of these
choices.  Finally, we developed a principled way of incorporating DA
modeling into the probability model of a continuous speech recognizer,
by constraining word hypotheses using the discourse context.  However,
the approach gives only a small reduction in word error on our corpus,
which can be attributed to a preponderance of a single dialogue act type
(statements).

\starttwocolumn

\small
\section*{Note}
\cbstart
The research described here is based on a project at the 1997 Workshop on
Innovative Techniques in LVCSR at the Center for Speech and Language
Processing at Johns Hopkins University
\cite{JurafskyEtAl:asru97,JurafskyEtAl:ws97}.
The DA-labeled Switchboard transcripts as well as other project-related
publications are available at
http://www.colorado.edu/\-ling/\-jurafsky/\-ws97/.
\cbend
\vspace{1em}

\begin{acknowledgments}

We thank the funders, researchers, and support staff of the
\cbstart
1997 Johns Hopkins Summer Workshop,
\cbend
especially Bill Byrne, Fred Jelinek,
Harriet Nock, Joe Picone, Kimberly Shiring, and Chuck Wooters.
Additional support came from the NSF via grants IRI-9619921 and IRI-9314967,
and from the UK Engineering and Physical Science Research Council (grant
GR/J55106).
Thanks to Mitch Weintraub, to Susann LuperFoy, Nigel Ward, James Allen,
Julia Hirschberg, and Marilyn
Walker for advice on the design of the SWBD-DAMSL tag-set,
to the discourse labelers at CU Boulder
(Debra Biasca, Marion Bond, Traci Curl, Anu Erringer, Michelle Gregory,
Lori Heintzelman, Taimi Metzler, and Amma Oduro) and the intonation
labelers at the University of Edinburgh
(Helen Wright, Kurt Dusterhoff, Rob Clark, Cassie Mayo, and Matthew Bull).
We also thank Andy Kehler and the anonymous reviewers for valuable
comments on a draft of this paper.
\end{acknowledgments}

\small
\bibliography{all}

\begin{thebibliography}{}

\bibitem[\protect\citename{Alexandersson and Reithinger}1997]{Alexandersson:97}
Alexandersson, Jan and Norbert Reithinger.
\newblock 1997.
\newblock Learning dialogue structures from a corpus.
\newblock In G.~Kokkinakis, N.~Fakotakis, and E.~Dermatas, editors, {\em
  Proceedings of the 5th European Conference on Speech Communication and
  Technology}, volume~4, pages 2231--2234, Rhodes, Greece, September.

\bibitem[\protect\citename{Anderson \bgroup et al.\egroup
  }1991]{AndersonEtAl:91}
Anderson, Anne~H., Miles Bader, Ellen~G. Bard, Elizabeth~H. Boyle, Gwyneth~M.
  Doherty, Simon~C. Garrod, Stephen~D. Isard, Jacqueline~C. Kowtko, Jan~M.
  McAllister, Jim Miller, Catherine~F. Sotillo, Henry~S. Thompson, and Regina
  Weinert.
\newblock 1991.
\newblock The {HCRC} {Map Task} corpus.
\newblock {\em Language and Speech}, 34(4):351--366.

\bibitem[\protect\citename{Austin}1962]{Austin:62}
Austin, J.~L.
\newblock 1962.
\newblock {\em How to do Things with Words}.
\newblock Clarendon Press, Oxford.

\bibitem[\protect\citename{Bahl, Jelinek, and Mercer}1983]{Bahl:83}
Bahl, Lalit~R., Frederick Jelinek, and Robert~L. Mercer.
\newblock 1983.
\newblock A maximum likelihood approach to continuous speech recognition.
\newblock {\em IEEE Transactions on Pattern Analysis and Machine Intelligence},
  5(2):179--190, March.

\bibitem[\protect\citename{Bard \bgroup et al.\egroup }1995]{BardEtAl:95}
Bard, Ellen~G., Catherine Sotillo, Anne~H. Anderson, and M.~M. Taylor.
\newblock 1995.
\newblock The {DCIEM} {Map Task} corpus: Spontaneous dialogues under sleep
  deprivation and drug treatment.
\newblock In Isabel Trancoso and Roger Moore, editors, {\em Proceedings of the
  {ESCA-NATO} Tutorial and Workshop on Speech under Stress}, pages 25--28,
  Lisbon, September.

\bibitem[\protect\citename{Baum \bgroup et al.\egroup }1970]{Baum:70}
Baum, Leonard~E., Ted Petrie, George Soules, and Norman Weiss.
\newblock 1970.
\newblock A maximization technique occurring in the statistical analysis of
  probabilistic functions in {Markov} chains.
\newblock {\em The Annals of Mathematical Statistics}, 41(1):164--171.

\bibitem[\protect\citename{Berger, {Della Pietra}, and {Della
  Pietra}}1996]{Berger:96}
Berger, Adam~L., Stephen~A. {Della Pietra}, and Vincent~J. {Della Pietra}.
\newblock 1996.
\newblock A maximum entropy approach to natural language processing.
\newblock {\em Computational Linguistics}, 22(1):39--71.

\bibitem[\protect\citename{Bourlard and Morgan}1993]{Bourlard:93}
Bourlard, Herv{\'e} and Nelson Morgan.
\newblock 1993.
\newblock {\em Connectionist Speech Recognition. A Hybrid Approach}.
\newblock Kluwer Academic Publishers, Boston, MA.

\bibitem[\protect\citename{Breiman \bgroup et al.\egroup }1984]{Breiman:84}
Breiman, L., J.~H. Friedman, R.~A. Olshen, and C.~J. Stone.
\newblock 1984.
\newblock {\em Classification and Regression Trees}.
\newblock Wadsworth and Brooks, Pacific Grove, CA.

\bibitem[\protect\citename{Bridle}1990]{Bridle:90}
Bridle, J.~S.
\newblock 1990.
\newblock Probabilistic interpretation of feedforward classification network
  outputs, with relationships to statistical pattern recognition.
\newblock In F.~Fogleman Soulie and J.~Herault, editors, {\em Neurocomputing:
  Algorithms, Architectures and Applications}. Springer, Berlin, pages
  227--236.

\bibitem[\protect\citename{Brill}1993]{Brill:93}
Brill, Eric.
\newblock 1993.
\newblock Automatic grammar induction and parsing free text: A
  transformation-based approach.
\newblock In {\em Proceedings of the {ARPA} Workshop on Human Language
  Technology}, Plainsboro, NJ, March.

\bibitem[\protect\citename{Carletta}1996]{Carletta:96}
Carletta, Jean.
\newblock 1996.
\newblock Assessing agreement on classification tasks: The {Kappa} statistic.
\newblock {\em Computational Linguistics}, 22(2):249--254.

\bibitem[\protect\citename{Carlson}1983]{Carlson:83}
Carlson, Lari.
\newblock 1983.
\newblock {\em Dialogue Games: An Approach to Discourse Analysis}.
\newblock D. Reidel.

\bibitem[\protect\citename{Chu-Carroll}1998]{Chu-Carroll:98}
Chu-Carroll, Jennifer.
\newblock 1998.
\newblock A statistical model for discourse act recognition in dialogue
  interactions.
\newblock In Jennifer Chu-Carroll and Nancy Green, editors, {\em Applying
  Machine Learning to Discourse Processing. Papers from the 1998 {AAAI} Spring
  Symposium. {\rm Technical Report SS-98-01}}, pages 12--17. AAAI Press, Menlo
  Park, CA.

\bibitem[\protect\citename{Church}1988]{Church:88}
Church, Kenneth~Ward.
\newblock 1988.
\newblock A stochastic parts program and noun phrase parser for unrestricted
  text.
\newblock In {\em Second Conference on Applied Natural Language Processing},
  pages 136--143, Austin, TX.

\bibitem[\protect\citename{Core and Allen}1997]{CoreAllen:97}
Core, Mark and James Allen.
\newblock 1997.
\newblock Coding dialogs with the {DAMSL} annotation scheme.
\newblock In {\em Working Notes of the {AAAI} Fall Symposium on Communicative
  Action in Humans and Machines}, pages 28--35, Cambridge, MA, November.

\bibitem[\protect\citename{Dermatas and Kokkinakis}1995]{Dermatas:95}
Dermatas, Evangelos and George Kokkinakis.
\newblock 1995.
\newblock Automatic stochastic tagging of natural language texts.
\newblock {\em Computational Linguistics}, 21(2):137--163.

\bibitem[\protect\citename{Eckert, Gallwitz, and Niemann}1996]{EckertEtAl:96}
Eckert, Wieland, Florian Gallwitz, and Heinrich Niemann.
\newblock 1996.
\newblock Combining stochastic and linguistic language models for recognition
  of spontaneous speech.
\newblock In {\em Proceedings of the IEEE Conference on Acoustics, Speech, and
  Signal Processing}, volume~1, pages 423--426, Atlanta, May.

\bibitem[\protect\citename{Finke \bgroup et al.\egroup }1998]{FinkeEtAl:aaai98}
Finke, Michael, Maria Lapata, Alon Lavie, Lori Levin, Laura~Mayfield Tomokiyo,
  Thomas Polzin, Klaus Ries, Alex Waibel, and Klaus Zechner.
\newblock 1998.
\newblock Clarity: Inferring discourse structure from speech.
\newblock In Jennifer Chu-Carroll and Nancy Green, editors, {\em Applying
  Machine Learning to Discourse Processing. Papers from the 1998 {AAAI} Spring
  Symposium. {\rm Technical Report SS-98-01}}, pages 25--32. AAAI Press, Menlo
  Park, CA.

\bibitem[\protect\citename{Fowler and Housum}1987]{FowlerHousum:87}
Fowler, Carol~A. and Jonathan Housum.
\newblock 1987.
\newblock Talkers' signaling of ``new'' and ``old'' words in speech and
  listeners' perception and use of the distinction.
\newblock {\em Journal of Memory and Language}, 26:489--504.

\bibitem[\protect\citename{Fukada \bgroup et al.\egroup }1998]{FukadaEtAl:98}
Fukada, Toshiaki, Detlef Koll, Alex Waibel, and Kouichi Tanigaki.
\newblock 1998.
\newblock Probabilistic dialogue act extraction for concept based multilingual
  translation systems.
\newblock In Robert~H. Mannell and Jordi Robert-Ribes, editors, {\em
  Proceedings of the International Conference on Spoken Language Processing},
  volume~6, pages 2771--2774, Sydney, December. Australian Speech Science and
  Technology Association.

\bibitem[\protect\citename{Godfrey, Holliman, and McDaniel}1992]{SWBD}
Godfrey, J.~J., E.~C. Holliman, and J.~McDaniel.
\newblock 1992.
\newblock {SWITCHBOARD}: {T}elephone speech corpus for research and
  development.
\newblock In {\em Proceedings of the IEEE Conference on Acoustics, Speech, and
  Signal Processing}, volume~1, pages 517--520, San Francisco, March.

\bibitem[\protect\citename{Grosz and Sidner}1986]{Grosz:86}
Grosz, B. and C.~Sidner.
\newblock 1986.
\newblock Attention, intention, and the structure of discourse.
\newblock {\em Computational Linguistics}, 12(3):175--204.

\bibitem[\protect\citename{Hirschberg and Litman}1993]{HirschbergLitman:93}
Hirschberg, Julia~B. and Diane~J. Litman.
\newblock 1993.
\newblock Empirical studies on the disambiguation of cue phrases.
\newblock {\em Computational Linguistics}, 19(3):501--530.

\bibitem[\protect\citename{Iyer, Ostendorf, and Rohlicek}1994]{Iyer:94}
Iyer, Rukmini, Mari Ostendorf, and J.~Robin Rohlicek.
\newblock 1994.
\newblock Language modeling with sentence-level mixtures.
\newblock In {\em Proceedings of the {ARPA} Workshop on Human Language
  Technology}, pages 82--86, Plainsboro, NJ, March.

\bibitem[\protect\citename{Jefferson}1984]{Jefferson:84}
Jefferson, Gail.
\newblock 1984.
\newblock Notes on a systematic deployment of the acknowledgement tokens `yeah'
  and `mm hm'.
\newblock {\em Papers in Linguistics}, 17:197--216.

\bibitem[\protect\citename{Jekat \bgroup et al.\egroup }1995]{JekatEtAl:95}
Jekat, Susanne, Alexandra Klein, Elisabeth Maier, Ilona Maleck, Marion Mast,
  and Joachim Quantz.
\newblock 1995.
\newblock Dialogue acts in {VERBMOBIL}.
\newblock Verbmobil-Report~65, Universit{\"a}t Hamburg, DFKI GmbH,
  Universit{\"ä}t Erlangen, and TU Berlin, April.

\bibitem[\protect\citename{Jurafsky \bgroup et al.\egroup
  }1997]{JurafskyEtAl:asru97}
Jurafsky, Dan, Rebecca Bates, Noah Coccaro, Rachel Martin, Marie Meteer, Klaus
  Ries, Elizabeth Shriberg, Andreas Stolcke, Paul Taylor, and Carol {Van
  Ess}-Dykema.
\newblock 1997.
\newblock Automatic detection of discourse structure for speech recognition and
  understanding.
\newblock In {\em Proceedings {IEEE} Workshop on Speech Recognition and
  Understanding}, pages 88--95, Santa Barbara, CA, December.

\bibitem[\protect\citename{Jurafsky \bgroup et al.\egroup
  }1998]{JurafskyEtAl:ws97}
Jurafsky, Daniel, Rebecca Bates, Noah Coccaro, Rachel Martin, Marie Meteer,
  Klaus Ries, Elizabeth Shriberg, Andreas Stolcke, Paul Taylor, and Carol {Van
  Ess}-Dykema.
\newblock 1998.
\newblock Switchboard discourse language modeling project final report.
\newblock Research Note~30, Center for Language and Speech Processing, Johns
  Hopkins University, Baltimore, January.

\bibitem[\protect\citename{Jurafsky, Shriberg, and
  Biasca}1997]{Jurafsky:97-damsl}
Jurafsky, Daniel, Elizabeth Shriberg, and Debra Biasca.
\newblock 1997.
\newblock Switchboard-{DAMSL} {L}abeling {P}roject {C}oder's {M}anual.
\newblock Technical Report 97-02, University of Colorado, Institute of
  Cognitive Science, Boulder, CO.
\newblock {\tt
  http://www.colorado.edu/\-ling/\-jurafsky/\-manual.august1.html}.

\bibitem[\protect\citename{Jurafsky \bgroup et al.\egroup }1998]{JurShrEtAl:98}
Jurafsky, Daniel, Elizabeth~E. Shriberg, Barbara Fox, and Traci Curl.
\newblock 1998.
\newblock Lexical, prosodic, and syntactic cues for dialog acts.
\newblock In {\em Proceedings of ACL/COLING-98 Workshop on Discourse Relations
  and Discourse Markers}, pages 114--120. Association for Computational
  Linguistics.

\bibitem[\protect\citename{Katz}1987]{Katz:87}
Katz, Slava~M.
\newblock 1987.
\newblock Estimation of probabilities from sparse data for the language model
  component of a speech recognizer.
\newblock {\em IEEE Transactions on Acoustics, Speech, and Signal Processing},
  35(3):400--401, March.

\bibitem[\protect\citename{Kita \bgroup et al.\egroup }1996]{KitaEtAl:96}
Kita, Kenji, Yoshikazu Fukui, Masaaki Nagata, and Tsuyoshi Morimoto.
\newblock 1996.
\newblock Automatic acquisition of probabilistic dialogue models.
\newblock In H.~Timothy Bunnell and William Idsardi, editors, {\em Proceedings
  of the International Conference on Spoken Language Processing}, volume~1,
  pages 196--199, Philadelphia, October.

\bibitem[\protect\citename{Kompe}1997]{Kompe:97}
Kompe, Ralf.
\newblock 1997.
\newblock {\em Prosody in speech understanding systems}.
\newblock Springer, Berlin.

\bibitem[\protect\citename{Kowtko}1996]{Kowtko:96}
Kowtko, Jacqueline~C.
\newblock 1996.
\newblock {\em The Function of Intonation in Task Oriented Dialogue}.
\newblock {Ph.D.} thesis, University of Edinburgh, Edinburgh.

\bibitem[\protect\citename{Kuhn and de Mori}1990]{KuhnDeMori:90}
Kuhn, Roland and Renato de~Mori.
\newblock 1990.
\newblock A cache-base natural language model for speech recognition.
\newblock {\em IEEE Transactions on Pattern Analysis and Machine Intelligence},
  12(6):570--583, June.

\bibitem[\protect\citename{Levin and Moore}1977]{LevinMoore:77}
Levin, Joan~A. and Johanna~A. Moore.
\newblock 1977.
\newblock Dialogue games: Metacommunication structures for natural language
  interaction.
\newblock {\em Cognitive Science}, 1(4):395--420.

\bibitem[\protect\citename{Levin \bgroup et al.\egroup }1999]{LevinEtAl:99}
Levin, Lori, Klaus Ries, Ann Thym\'{e}-Gobbel, and Alon Lavie.
\newblock 1999.
\newblock Tagging of speech acts and dialogue games in {Spanish} {CallHome}.
\newblock In {\em Towards Standards and Tools for Discourse Tagging {\rm
  (Proceedings of the Workshop at ACL'99)}}, pages 42--47, College Park, MD,
  June.

\bibitem[\protect\citename{Linell}1990]{Linell:90}
Linell, Per.
\newblock 1990.
\newblock The power of dialogue dynamics.
\newblock In Ivana Markov{\'a} and Klaus Foppa, editors, {\em The Dynamics of
  Dialogue}. Harvester, Wheatsheaf, New York, London, pages 147--177.

\bibitem[\protect\citename{Mast \bgroup et al.\egroup }1996]{MastEtAl:96}
Mast, M., R.~Kompe, S.~Harbeck, A.~Kie{\ss}ling, H.~Niemann, E.~N{\"o}th, E.~G.
  Schukat-Talamazzini, and V.~Warnke.
\newblock 1996.
\newblock Dialog act classification with the help of prosody.
\newblock In H.~Timothy Bunnell and William Idsardi, editors, {\em Proceedings
  of the International Conference on Spoken Language Processing}, volume~3,
  pages 1732--1735, Philadelphia, October.

\bibitem[\protect\citename{Menn and Boyce}1982]{MennBoyce:82}
Menn, Lise and Suzanne~E. Boyce.
\newblock 1982.
\newblock Fundamental frequency and discourse structure.
\newblock {\em Language and Speech}, 25:341--383.

\bibitem[\protect\citename{Meteer \bgroup et al.\egroup }1995]{SWBD-DF}
Meteer, Marie, Ann Taylor, Robert MacIntyre, and Rukmini Iyer.
\newblock 1995.
\newblock Dysfluency annotation stylebook for the {Switchboard} corpus.
\newblock Distributed by LDC, {\tt
  ftp://ftp.cis.upenn.edu\-/pub\-/treebank\-/swbd\-/doc\-/DFL-book.ps},
  February.
\newblock Revised June 1995 by Ann Taylor.

\bibitem[\protect\citename{Morgan, Fosler, and
  Mirghafori}1997]{Morgan-enrate:97}
Morgan, Nelson, Eric Fosler, and Nikki Mirghafori.
\newblock 1997.
\newblock Speech recognition using on-line estimation of speaking rate.
\newblock In G.~Kokkinakis, N.~Fakotakis, and E.~Dermatas, editors, {\em
  Proceedings of the 5th European Conference on Speech Communication and
  Technology}, volume~4, pages 2079--2082, Rhodes, Greece, September.

\bibitem[\protect\citename{Munk}1999]{Munk:99}
Munk, Marcus.
\newblock 1999.
\newblock {\em Shallow Statistical Parsing for Machine Translation}.
\newblock Diploma thesis, Carnegie Mellon University.

\bibitem[\protect\citename{Nagata}1992]{Nagata:92}
Nagata, Masaaki.
\newblock 1992.
\newblock Using pragmatics to rule out recognition errors in cooperative
  task-oriented dialogues.
\newblock In John~J. Ohala, Terrance~M. Nearey, Bruce~L. Derwing, Megan~M.
  Hodge, and Grace~E. Wiebe, editors, {\em Proceedings of the International
  Conference on Spoken Language Processing}, volume~1, pages 647--650, Banff,
  Canada, October.

\bibitem[\protect\citename{Nagata and Morimoto}1993]{NagataMorimoto:93}
Nagata, Masaaki and Tsuyoshi Morimoto.
\newblock 1993.
\newblock An experimental statistical dialogue model to predict the speech act
  type of the next utterance.
\newblock In Katsuhiko Shirai, Tetsunori Kobayashi, and Yasunari Harada,
  editors, {\em Proceedings of the International Symposium on Spoken Dialogue},
  pages 83--86, Tokyo, November.

\bibitem[\protect\citename{Nagata and Morimoto}1994]{NagataMorimoto:94}
Nagata, Masaaki and Tsuyoshi Morimoto.
\newblock 1994.
\newblock First steps toward statistical modeling of dialogue to predict the
  speech act type of the next utterance.
\newblock {\em Speech Communication}, 15:193--203.

\bibitem[\protect\citename{Ohler, Harbeck, and Niemann}1999]{OhlerEtAl:99}
Ohler, Uwe, Stefan Harbeck, and Heinrich Niemann.
\newblock 1999.
\newblock Discriminative training of language model classifiers.
\newblock In {\em Proceedings of the 6th European Conference on Speech
  Communication and Technology}, volume~4, pages 1607--1610, Budapest,
  September.

\bibitem[\protect\citename{Pearl}1988]{Pearl:88}
Pearl, Judea.
\newblock 1988.
\newblock {\em Probabilistic Reasoning in Intelligent Systems: Networks of
  Plausible Inference}.
\newblock Morgan Kaufmann, San Mateo, CA.

\bibitem[\protect\citename{Power}1979]{Power:79}
Power, Richard J.~D.
\newblock 1979.
\newblock The organisation of purposeful dialogues.
\newblock {\em Linguistics}, 17:107--152.

\bibitem[\protect\citename{Rabiner and Juang}1986]{Rabiner:86}
Rabiner, L.~R. and B.~H. Juang.
\newblock 1986.
\newblock An introduction to hidden {Markov} models.
\newblock {\em IEEE ASSP Magazine}, 3(1):4--16, January.

\bibitem[\protect\citename{Reithinger \bgroup et al.\egroup
  }1996]{Reithinger:96}
Reithinger, Norbert, Ralf Engel, Michael Kipp, and Martin Klesen.
\newblock 1996.
\newblock Predicting dialogue acts for a speech-to-speech translation system.
\newblock In H.~Timothy Bunnell and William Idsardi, editors, {\em Proceedings
  of the International Conference on Spoken Language Processing}, volume~2,
  pages 654--657, Philadelphia, October.

\bibitem[\protect\citename{Reithinger and Klesen}1997]{Reithinger:97}
Reithinger, Norbert and Martin Klesen.
\newblock 1997.
\newblock Dialogue act classification using language models.
\newblock In G.~Kokkinakis, N.~Fakotakis, and E.~Dermatas, editors, {\em
  Proceedings of the 5th European Conference on Speech Communication and
  Technology}, volume~4, pages 2235--2238, Rhodes, Greece, September.

\bibitem[\protect\citename{Ries}1999a]{Ries:99}
Ries, Klaus.
\newblock 1999a.
\newblock {HMM} and neural network based speech act classification.
\newblock In {\em Proceedings of the IEEE Conference on Acoustics, Speech, and
  Signal Processing}, volume~1, pages 497--500, Phoenix, AZ, March.

\bibitem[\protect\citename{Ries}1999b]{Ries:99b}
Ries, Klaus.
\newblock 1999b.
\newblock Towards the detection and description of textual meaning indicators
  in spontaneous conversations.
\newblock In {\em Proceedings of the 6th European Conference on Speech
  Communication and Technology}, volume~3, pages 1415--1418, Budapest,
  September.

\bibitem[\protect\citename{Rose, Chang, and Lippmann}1991]{Rose:91}
Rose, R.~C., E.~I. Chang, and R.~P. Lippmann.
\newblock 1991.
\newblock Techniques for information retrieval from voice messages.
\newblock In {\em Proceedings of the IEEE Conference on Acoustics, Speech, and
  Signal Processing}, volume~1, pages 317--320, Toronto, May.

\bibitem[\protect\citename{Sacks, Schegloff, and Jefferson}1974]{Sacks:74}
Sacks, H., E.~A. Schegloff, and G.~Jefferson.
\newblock 1974.
\newblock A simplest semantics for the organization of turn-taking in
  conversation.
\newblock {\em Language}, 50(4):696--735.

\bibitem[\protect\citename{Samuel, Carberry, and Vijay-Shanker}1998]{Samuel:98}
Samuel, Ken, Sandra Carberry, and K.~Vijay-Shanker.
\newblock 1998.
\newblock Dialogue act tagging with transformation-based learning.
\newblock In {\em Proceedings of the 36th Annual Meeting of the Association for
  Computational Linguistics and 17th International Conference on Computational
  Linguistics}, volume~2, pages 1150--1156, Montreal.

\bibitem[\protect\citename{Schegloff}1968]{Schegloff:68}
Schegloff, E.~A.
\newblock 1968.
\newblock Sequencing in conversational openings.
\newblock {\em American Anthropologist}, 70:1075--1095.

\bibitem[\protect\citename{Schegloff}1982]{Schegloff:82}
Schegloff, Emanuel~A.
\newblock 1982.
\newblock Discourse as an interactional achievement: {S}ome uses of `uh huh'
  and other things that come between sentences.
\newblock In Deborah Tannen, editor, {\em Analyzing Discourse: {T}ext and
  Talk}. Georgetown University Press, Washington, D.C., pages 71--93.

\bibitem[\protect\citename{Searle}1969]{Searle:69}
Searle, J.~R.
\newblock 1969.
\newblock {\em Speech Acts}.
\newblock Cambridge University Press, London-New York.

\bibitem[\protect\citename{Shriberg \bgroup et al.\egroup
  }1998]{ShribergEtAl:LS}
Shriberg, Elizabeth, Rebecca Bates, Andreas Stolcke, Paul Taylor, Daniel
  Jurafsky, Klaus Ries, Noah Coccaro, Rachel Martin, Marie Meteer, and Carol
  {Van Ess-Dykema}.
\newblock 1998.
\newblock Can prosody aid the automatic classification of dialog acts in
  conversational speech?
\newblock {\em Language and Speech}, 41(3-4):439--487.

\bibitem[\protect\citename{Shriberg \bgroup et al.\egroup
  }2000]{ShribergEtAl:specom2000}
Shriberg, Elizabeth, Andreas Stolcke, Dilek Hakkani-T{\"u}r, and G{\"o}khan
  T{\"u}r.
\newblock 2000.
\newblock Prosody-based automatic segmentation of speech into sentences and
  topics.
\newblock {\em Speech Communication}, 32(1-2):to appear, September.
\newblock Special Issue on Accessing Information in Spoken Audio.

\bibitem[\protect\citename{Siegel and {Castellan, Jr.}}1988]{Siegel:88}
Siegel, Sidney and N.~John {Castellan, Jr.}
\newblock 1988.
\newblock {\em Nonparametric Statistics for the Behavioral Sciences}.
\newblock McGraw-Hill, New York, second edition.

\bibitem[\protect\citename{Stolcke and Shriberg}1996]{StoShr:icslp96}
Stolcke, Andreas and Elizabeth Shriberg.
\newblock 1996.
\newblock Automatic linguistic segmentation of conversational speech.
\newblock In H.~Timothy Bunnell and William Idsardi, editors, {\em Proceedings
  of the International Conference on Spoken Language Processing}, volume~2,
  pages 1005--1008, Philadelphia, October.

\bibitem[\protect\citename{Suhm and Waibel}1994]{SuhmWaibel:94}
Suhm, B. and A.~Waibel.
\newblock 1994.
\newblock Toward better language models for spontaneous speech.
\newblock In {\em Proceedings of the International Conference on Spoken
  Language Processing}, volume~2, pages 831--834, Yokohama, September.

\bibitem[\protect\citename{Taylor \bgroup et al.\egroup }1998]{TaylorEtAl:LS98}
Taylor, Paul, Simon King, Stephen Isard, and Helen Wright.
\newblock 1998.
\newblock Intonation and dialog context as constraints for speech recognition.
\newblock {\em Language and Speech}, 41(3-4):489--508.

\bibitem[\protect\citename{Taylor \bgroup et al.\egroup }1997]{Taylor:97}
Taylor, Paul, Simon King, Stephen Isard, Helen Wright, and Jacqueline Kowtko.
\newblock 1997.
\newblock Using intonation to constrain language models in speech recognition.
\newblock In G.~Kokkinakis, N.~Fakotakis, and E.~Dermatas, editors, {\em
  Proceedings of the 5th European Conference on Speech Communication and
  Technology}, volume~5, pages 2763--2766, Rhodes, Greece, September.

\bibitem[\protect\citename{Taylor}2000]{Taylor:2000}
Taylor, Paul~A.
\newblock 2000.
\newblock Analysis and synthesis of intonation using the tilt model.
\newblock {\em Journal of the Acoustical Society of America},
  107(3):1697--1714.

\bibitem[\protect\citename{{Van Ess}-Dykema and Ries}1998]{VanEss:98}
{Van Ess}-Dykema, Carol and Klaus Ries.
\newblock 1998.
\newblock Linguistically engineered tools for speech recognition error
  analysis.
\newblock In Robert~H. Mannell and Jordi Robert-Ribes, editors, {\em
  Proceedings of the International Conference on Spoken Language Processing},
  volume~5, pages 2091--2094, Sydney, December. Australian Speech Science and
  Technology Association.

\bibitem[\protect\citename{Viterbi}1967]{Viterbi:67}
Viterbi, A.
\newblock 1967.
\newblock Error bounds for convolutional codes and an asymptotically optimum
  decoding algorithm.
\newblock {\em IEEE Transactions on Information Theory}, 13:260--269.

\bibitem[\protect\citename{Warnke \bgroup et al.\egroup }1997]{WarnkeEtAl:97}
Warnke, V., R.~Kompe, H.~Niemann, and E.~N{\"o}th.
\newblock 1997.
\newblock Integrated dialog act segmentation and classification using prosodic
  features and language models.
\newblock In G.~Kokkinakis, N.~Fakotakis, and E.~Dermatas, editors, {\em
  Proceedings of the 5th European Conference on Speech Communication and
  Technology}, volume~1, pages 207--210, Rhodes, Greece, September.

\bibitem[\protect\citename{Warnke \bgroup et al.\egroup }1999]{WarnkeEtAl:99}
Warnke, Volker, Stefan Harbeck, Elmar N{\"o}th, Heinrich Niemann, and Michael
  Levit.
\newblock 1999.
\newblock Discriminative estimation of interpolation parameters for language
  model classifiers.
\newblock In {\em Proceedings of the IEEE Conference on Acoustics, Speech, and
  Signal Processing}, volume~1, pages 525--528, Phoenix, AZ, March.

\bibitem[\protect\citename{Weber}1993]{Weber:93}
Weber, Elizabeth~G.
\newblock 1993.
\newblock {\em Varieties of Questions in {E}nglish Conversation}.
\newblock John Benjamins, Amsterdam.

\bibitem[\protect\citename{Witten and Bell}1991]{Witten:91}
Witten, Ian~H. and Timothy~C. Bell.
\newblock 1991.
\newblock The zero-frequency problem: Estimating the probabilities of novel
  events in adaptive text compression.
\newblock {\em IEEE Transactions on Information Theory}, 37(4):1085--1094,
  July.

\bibitem[\protect\citename{Woszczyna and Waibel}1994]{WoszczynaWaibel:94}
Woszczyna, M. and A.~Waibel.
\newblock 1994.
\newblock Inferring linguistic structure in spoken language.
\newblock In {\em Proceedings of the International Conference on Spoken
  Language Processing}, volume~2, pages 847--850, Yokohama, September.

\bibitem[\protect\citename{Wright}1998]{Wright:98}
Wright, Helen.
\newblock 1998.
\newblock Automatic utterance type detection using suprasegmental features.
\newblock In Robert~H. Mannell and Jordi Robert-Ribes, editors, {\em
  Proceedings of the International Conference on Spoken Language Processing},
  volume~4, pages 1403--1406, Sydney, December. Australian Speech Science and
  Technology Association.

\bibitem[\protect\citename{Wright and Taylor}1997]{WrightTaylor:97}
Wright, Helen and Paul~A. Taylor.
\newblock 1997.
\newblock Modelling intonational structure using hidden {Markov} models.
\newblock In {\em Intonation: Theory, Models and Applications. Proceedings of
  an {ESCA} Workshop}, pages 333--336, Athens, September.

\bibitem[\protect\citename{Yngve}1970]{Yngve:70}
Yngve, Victor~H.
\newblock 1970.
\newblock On getting a word in edgewise.
\newblock In {\em Papers from the Sixth Regional Meeting of the Chicago
  Linguistic Society}, pages 567--577, Chicago, April. University of Chicago.

\bibitem[\protect\citename{Yoshimura \bgroup et al.\egroup
  }1996]{YoshimuraEtAl:96}
Yoshimura, Takashi, Satoru Hayamizu, Hiroshi Ohmura, and Kazuyo Tanaka.
\newblock 1996.
\newblock Pitch pattern clustering of user utterances in human-machine
  dialogue.
\newblock In H.~Timothy Bunnell and William Idsardi, editors, {\em Proceedings
  of the International Conference on Spoken Language Processing}, volume~2,
  pages 837--840, Philadelphia, October.

\end{thebibliography}

\end{document}